\documentclass{article} 
\PassOptionsToPackage{numbers,sort,compress}{natbib}
\usepackage[preprint]{neurips_2022}

\usepackage[utf8]{inputenc} 
\usepackage[T1]{fontenc}    
\usepackage{url}            
\usepackage{booktabs}       
\usepackage{amsfonts}       
\usepackage{nicefrac}       
\usepackage{microtype}      
\usepackage{xcolor}         
\usepackage{graphicx}
\usepackage{tikz}
\usepackage{comment}
\usepackage[english]{babel}
\usepackage{amsthm,amsmath,amsfonts,amssymb,mathrsfs}
\usepackage{enumitem}
\usepackage[hang,marginal]{footmisc} 
\usepackage[percent]{overpic} 
\usepackage{mathabx} 
\usepackage{bbm}
\usepackage{color,soul}
\usepackage{makecell}
\usepackage{tabularx,ragged2e}
\newcolumntype{Y}{>{\centering\arraybackslash}X}
\usepackage{booktabs}
\usepackage{mwe}
\usepackage{graphbox}
\usepackage[font={small}]{caption}
\usepackage{subcaption}
\usepackage{wrapfig}
\usepackage[pagebackref=true,breaklinks=true,colorlinks,bookmarks=false]{hyperref}
\usepackage[framemethod=TikZ]{mdframed}
\usepackage{floatrow}
\usepackage{setspace}
\theoremstyle{definition}
\newtheorem{definition}{Definition}
\newtheorem{proposition}{Proposition}

\mdfdefinestyle{yellowFrame}{%
  linecolor=midGray,
  linewidth=0.5pt,
  roundcorner=3pt,
  innertopmargin=5pt,
  innerbottommargin=5pt,
  innerrightmargin=6pt,
  innerleftmargin=6pt,
  leftmargin=0pt,
  rightmargin=0pt,
  backgroundcolor=lightYellow
}

\mdfdefinestyle{grayFrame}{%
  linecolor=midGray,
  linewidth=0.5pt,
  roundcorner=3pt,
  innertopmargin=6pt,
  innerbottommargin=6pt,
  innerrightmargin=6pt,
  innerleftmargin=6pt,
  leftmargin=0pt,
  rightmargin=0pt,
  backgroundcolor=lightGray
}

\mdfdefinestyle{codeFrame}{%
  linecolor=black,
  linewidth=0.5pt,
  innertopmargin=5pt,
  innerbottommargin=5pt,
  innerrightmargin=6pt,
  innerleftmargin=6pt,
  leftmargin=0pt,
  rightmargin=0pt
}

\def\onedot{\ifx\@let@token.\else.\null\fi\xspace}
\def\eg{\emph{e.g}\onedot} 
\def\ie{\emph{i.e}\onedot} 
 
 \def\vs{\emph{vs}\onedot}
\def\etal{\emph{et al}\onedot}

\usepackage[noend,ruled]{algorithm2e}
\SetKwInput{KwInput}{Inputs}
\SetKwInput{KwOutput}{Result}
\SetKwInput{KwResult}{Method}
\SetKwRepeat{Do}{do}{while}
\newcommand\myComment[1]{\textnormal{\textcolor{midGray}{//~\footnotesize#1}}}
\SetCommentSty{myComment}

\definecolor{midGray}{gray}{0.55}
\definecolor{lightGray}{gray}{0.96}
\definecolor{darkBlue}{rgb}{0,0,0.75}
\definecolor{darkGreen}{rgb}{0,0.75,0}
\definecolor{lightYellow}{RGB}{252,252,240}

\usepackage{listings}
\usepackage{color}

\newcommand{\bb}{\boldsymbol{b}}

\newcommand{\bv}{{\boldsymbol{v}}}
\newcommand{\bw}{{\boldsymbol{w}}}
\newcommand{\bx}{{\boldsymbol{x}}}

\newcommand{\bI}{{\boldsymbol{I}}}

\newcommand{\bS}{{\boldsymbol{S}}}

\newcommand{\bW}{{\boldsymbol{W}}}

\newcommand{\btheta}{{\boldsymbol{\theta}}}

\newcommand{\bpsi}{{\boldsymbol{\psi}}}

\newcommand{\figref}[1]{Figure~\ref{#1}}

\newcommand{\secref}[1]{Section~\ref{#1}}
\newcommand{\appref}[1]{Appendix~\ref{#1}}
\newcommand{\tabref}[1]{Table~\ref{#1}}
\newcommand{\algref}[1]{Algorithm~\ref{#1}}
\newcommand{\defref}[1]{Definition~\ref{#1}}

\newcommand{\pp}[1]{{\tiny{$\pm${#1}}}}
\newcommand{\hangBoxC}[1]{%
  \begin{minipage}[c]{\linewidth}\begin{tabbing} 
  ~\\[-\baselineskip] 
  #1 
  \end{tabbing}
  \end{minipage}}

\newcommand{\myparagraphWoSpacing}[1]{\vspace{-1pt}\noindent{\normalsize\bfseries #1}\hspace{.05em}} 
\newcommand{\myparagraph}[1]{\vspace{-1pt}\noindent{\normalsize\bfseries #1}\hspace{.03em}} 

\newcommand{\minSpace}{\hspace{-.5em}}
\newcommand{\minSpaceSmall}{\hspace{-.15em}}

\newcommand{\smallEq}{\hspace{-.18em}=\hspace{-.18em}}
\newcommand{\smallNeq}{\hspace{-.18em}\neq\hspace{-.18em}}
\newcommand{\smallApprox}{\hspace{-.18em}\approx\hspace{-.18em}}
\newcommand{\smallSim}{\hspace{-.18em}\sim\hspace{-.18em}}

\newcommand{\aenc}{\mathrm{proj}}
\newcommand{\din}{{d_{\textrm{in}}}}
\newcommand{\dman}{{d_{\textrm{manifold}}}}
\newcommand{\manifold}{\mathcal{M}}
\newcommand{\lossPred}{\mathcal{L}_\textrm{pred}}
\newcommand{\lossIndep}{\mathcal{L}_\textrm{indep}}
\newcommand{\lossOnManifold}{\mathcal{L}_\textrm{manifold}}
\newcommand{\wtIndep}{{\lambda_\textrm{indep}}}
\newcommand{\wtManifold}{{\lambda_\textrm{manifold}}}
\newcommand{\data}{\mathcal{D}}
\newcommand{\dataTr}{\mathcal{D}_\mathrm{tr}}
\newcommand{\dataVal}{\mathcal{D}_\mathrm{val}}
\newcommand{\dataTest}{\mathcal{D}_\mathrm{test}}
\newcommand{\dataUnlabelled}{\mathcal{D}_\mathrm{OOD}}
\newcommand{\deltax}{\bv}

\newcommand{\mask}{{\boldsymbol{\mathrm{mask}}}}
\newcommand{\argmin}{\mathop{\textrm{argmin}}}
\newcommand{\argmax}{\mathop{\textrm{argmax}}}
\newcommand{\vol}{\mathop{\textrm{vol}}}
\DeclareMathOperator*{\prob}{\textrm{P}} 
\DeclareMathOperator*{\mi}{\textrm{MI}}
\DeclareMathOperator*{\sig}{\sigma}
\newcommand{\project}{\mathrm{proj}}
\DeclareMathOperator*{\indicator}{\mathbbm{1}\hspace{-.2em}} 
\DeclareMathOperator*{\indicatorSmall}{\mathbbm{1}\hspace{-.1em}} 

\DeclareMathOperator*{\risk}{\mathcal{R}}
\newcommand*{\mydots}{\kern-0.10em.\kern-0.10em.\kern-0.10em.} 
\newcommand{\indep}{\perp}
\newcommand{\probId}{\textrm{P}_{\mathrm{ID}}}
\newcommand{\probOod}{\textrm{P}_{\mathrm{OOD}}}

\title{Predicting is not Understanding:\\Recognizing and Addressing Underspecification\\in Machine Learning}

\author{Damien Teney$^{1,\!3}$ \quad Maxime Peyrard$^2$ \quad Ehsan Abbasnejad$^3$\vspace{2pt}\\
	$^1$Idiap Research Institute \quad $^2$EPFL \quad $^3$Australian Institute for Machine Learning}

\begin{document}

\maketitle

\vspace{-16pt}
\begin{abstract}
Machine learning (ML) models are typically optimized for their accuracy on a given dataset.
However, this predictive criterion rarely captures all desirable properties of a model, in particular how well it matches a domain expert's \emph{understanding} of a task.
Underspecification~\cite{d2020underspecification} refers to the existence of multiple models that are indistinguishable in their in-domain accuracy, even though they differ in other desirable properties such as out-of-distribution (OOD) performance.
Identifying these situations is critical for assessing the reliability of ML models.

\vspace{1pt}
We formalize the concept of underspecification and propose a method to identify and partially address it.
We train multiple models with an independence constraint that forces them to implement different functions.
They discover predictive features that are otherwise ignored by standard empirical risk minimization (ERM), which we then distill into a global model with superior OOD performance.
Importantly, we constrain the models to align with the data manifold to ensure that they discover meaningful features.
We demonstrate the method on multiple datasets in computer vision (collages, WILDS-Camelyon17, GQA) and discuss general implications of underspecification.
Most notably, in-domain performance cannot serve for OOD model selection without additional assumptions.
\end{abstract}

\section{Introduction}
\label{secIntro}
\myparagraph{Is {data} all you need?}
A finite set of i.i.d. examples is almost never sufficient to learn a task.
Inductive biases have long been known to be necessary for {in-domain} generalization~\cite{mitchell1980need,wolpert1996lack}.
Out-of-distribution~(OOD)%
\footnote{In this paper, OOD refers to test data that has undergone a covariate shift~\cite{shimodaira2000improving} compared to the training data.}
generalization complicates things further since one also needs to determine which predictive patterns of the training data will remain relevant at test time.
Correlations between inputs and labels that are
important for the task may be indistinguishable from spurious ones that result from dataset-specific artefacts such as selection biases.

\myparagraph{An example in image recognition.} Image labels are often correlated with objects and the backgrounds they appear in (\eg cars in cities, birds in nature).
Recognizing either often suffice to predict correct labels.
However, robust OOD generalization (\eg correctly labeling images of birds in street scenes) requires to rely on shapes and to ignore the background.
When this requirement cannot be deduced from the data
(because both features leave a similar signature in the joint training distribution),
the task is said to be underspecified.
In this example, the task requires the additional knowledge that labels refer to object shapes rather than background textures~\cite{geirhos2018imagenet}.
Such knowledge is often task-specific.
For example, the opposite choice of prioritizing color or texture over shape would be sensible for recognizing traffic signs or segmenting medical images.

\vspace{8pt}
\begin{mdframed}[style=yellowFrame,userdefinedwidth=.77\linewidth,align=center]
  \textbf{Underspecification gap}:
  the difference between the information provided in a dataset and the information required to perform \emph{as desired} on a task.
\end{mdframed}
\vspace{1pt}
\clearpage

\noindent
The qualifier ``\emph{as desired}'' captures the fact that different use cases require different properties such as adversarial robustness, interpretability, fairness, or OOD generalization. The latter is the focus of this paper.
Underspecification arises because these properties do not necessarily correlate with the ERM objective~\cite{vapnik1998statistical} typically used to train models.

This paper argues that \textbf{identifying underspecification}
is important for
assessing the reliability of ML models,
their reliance on hidden assumptions,
and for identifying the information missing for OOD robustness.
We identify underspecification by \textbf{discovering~multiple~understandings} of the data.
We learn multiple predictive models compatible with a given dataset and hypothesis class (low in-domain risk).
We force them to rely on different predictive features by encouraging orthogonality of their input gradients. We also ensure that these features remain semantically meaningful by constraining the input gradients to the data manifold.
Training multiple models stands in contrast with the standard practice of optimizing a single solution to a learning problem~--~which hides the existence of underspecification.
With our method, we discover predictive features otherwise ignored by standard ERM.
This alone produces candidate models with superior OOD performance.
In addition, we show how to distill selected features from multiple candidate models into one that is robust across a wider range of distribution shifts.
In all cases, a selection strategy must be provided  (see \secref{secSelectingFeatures}) such as an OOD validation set, domain expertise, task-specific heuristics, etc.

\begin{figure}[t!]
  \centering
  \includegraphics[width=\linewidth]{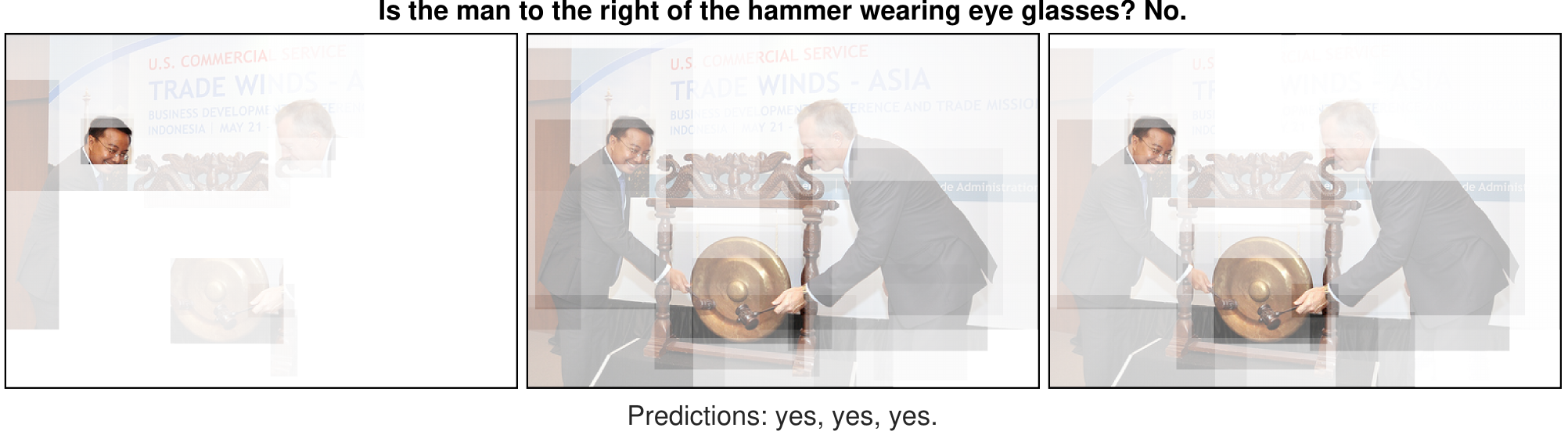}
  \vspace{-2pt}
  \caption{
  Example of underspecification in visual question answering.
  Our method trains multiple models that each discover different predictive features.
  We obtain three models producing identical answers on most training and validation data, even though they rely on different visual clues (evidenced by grad-CAM visualizations over object proposals \cite{selvaraju2019taking}).
  Each model reflects a \textbf{different understanding of the task} compatible with the data (possibly incomprehensible to humans) which reveals ambiguity in its specification.
  \label{figTeaser}}
\end{figure}

\myparagraph{Experiments.}
We apply the method to controlled data (collages~\cite{shah2020pitfalls,teney2021evading}) and computer vision benchmarks (WILDS-Camelyon17~\cite{koh2021wilds}, GQA~\cite{hudson2018gqa,kervadec2021roses}).
On visual question answering, we show that multiple models can produce similar answers while relying on different visual features~(\figref{figTeaser}).

\myparagraph{Implications.}
\looseness=-1
Our work complements other studies~\cite{d2020underspecification,mehrer2020individual} in formalizing underspecification as a root cause of multiple challenges in ML including shortcut learning, distribution shifts, and even adversarial vulnerabilities (an extreme case of OOD inputs).
Our formalization of underspecification makes it obvious that ID and OOD performance are not necessarily coupled.
Therefore, without further assumptions, {in-domain validation performance is not a reliable model selection strategy for OOD performance} despite contradictory suggestions made in the literature~\cite{gulrajani2020search,miller2021accuracy}.
The prevalence of underspecification~\cite{d2020underspecification} also suggests that {task-specific knowledge and assumptions are often necessary to build robust ML models, since they cannot emerge from simply scaling up data and architectures}.

\vspace{2pt}
\myparagraph{Summary of contributions.}
\setlist{nolistsep,leftmargin=*}
\begin{enumerate}[topsep=-2pt,itemsep=2pt]
  \item We propose a mathematical framework for quantifying and addressing underspecification.
  \item We derive a method to learn a set of models compatible with a given dataset that exhibit distinct OOD behaviour. We force the models to rely on different features (independence objective) that are nonetheless semantically meaningful (on-manifold constraint).
  \item We use the method for (1)~highlighting underspecification in given dataset/architecture pairs, and (2)~building models with superior OOD performance. We use controlled and real-world datasets (collages~\cite{shah2020pitfalls,teney2021evading}, WILDS-Camelyon17~\cite{koh2021wilds}, GQA~\cite{hudson2018gqa,kervadec2021roses}).
\end{enumerate}

\clearpage

\section{Related work}
\label{secRelatedWork}

\myparagraphWoSpacing{Underspecification in ML.}
An empirical study by d'Amour \etal in~\cite{d2020underspecification} showed that models trained with different random seeds have wide variations in OOD performance despite similar in-domain performance.
The effect was surprising in its prevalence but it echoed earlier observations~\cite{mehrer2020individual}.
It results from the well-known impossibility of achieving OOD generalization solely through ERM and i.i.d. data~\cite{arjovsky2019invariant,scholkopf2021toward}.
Indeed, since this learning objective does not constrain the model's behaviour outside the training distribution (\ie multiple hypotheses are plausible), additional assumptions/expertise is necessary \eg to tweak the network architecture to the task.
Data-driven OOD generalization is also possible but it requires heterogeneous (non-i.i.d.) data such as multiple environments~\cite{arjovsky2019invariant,peters2016causal,teney2020unshuffling}, counterfactual examples~\cite{heinze2017conditional,teney2021evading}, or non-stationary data~\cite{halva2020hidden,hyvarinen2017nonlinear,pfister2019invariant}.
The impossibility of OOD generalization from i.i.d. data is intimately related to identifiability in causal discovery~\cite{bareinboim2020pearl} and non-linear ICA~\cite{khemakhem2020variational}, and the impossibility of unsupervised disentanglement in representation learning~\cite{locatello2019challenging}.

\paragraph{Multiple hypotheses compatible with the data.}
The \textbf{Rashomon effect}~\cite{breiman2001statistical} is almost synonymous with underspecfication but it is agnostic to OOD data.
The Rashomon {set}~\cite{fisher2019all}
is the set of models of a given class whose training loss or in-domain (ID) risk lies below a threshold~\cite{hsu2022rashomon,rudin2021interpretable,semenova2019study}.
\textbf{Predictive multiplicity}~\cite{bhatt2020counterfactual,marx2020predictive,renard2021understanding} refers to the existence of multiple models compatible with the data that make conflicting predictions (on ID data again, rather than OOD).
This paper is about models with low ID risk,
yet different OOD behaviour.
We propose a method to~\textbf{learn~multiple~models} compatible with the data.
Related approaches were described for ensembling~\cite{ross2020ensembles}, interpretability~\cite{ross2018learning}, and control of inductive biases~\cite{teney2021evading}.
The number of models required in~\cite{teney2021evading} was very large.
We solve this issue by constraining models to align with the data manifold.
Other works concerned with the identification of multiple solutions to a learning problem include~\cite{parker2020ridge}, Bayesian deep learning~\cite{wang2020survey}, and classical feature selection~\cite{guyon2003introduction},
none of which are suitable to large-scale models and datasets.

\paragraph{Ensembles.}
Our method for combining robust features resembles traditional ensembling, which lowers variance by combining predictors with uncorrelated errors.
In comparison however, we lower prediction bias by selecting features causally related to the target.
The key in our approach is to discover predictive features that are otherwise missed by ERM.

\paragraph{Diversity in feature space.}
Allen-Zhu and Li~\cite{allen2020towards} explained the success of deep ensembles with the use of different features by different networks simply because of different initializations.
Their argument is also based on the ubiquity of underspecification which they call ``multi-view structure'' of data.
Concurrently to our work, \cite{yashima2022feature}~used this explanation to build better ensembles by encouraging diversity in \emph{feature space} with a Bayesian framework.
Even more recently, Heljakka \etal~\cite{heljakka2022representational} showed how to quantify representational (\ie feature-wise) multiplicity.

\paragraph{Diversity in prediction space.}
Concurrently to our work, \cite{pagliardini2022agree} and \cite{lee2022diversify}
proposed to learn {models that maximally disagree} in OOD predictions.
Both learn only two models.
Obviously, neither would do well on our \textit{collage} dataset (\secref{secCollages}) which requires discovering 4 predictive signals.
On real vision datasets, our experiments also suggest that many predictive features ($\gg$2) can be discovered.

\section{Formalizing underspecification}
\label{secDefinitions}

\looseness=-1
Let us focus on binary classification tasks.
A {dataset} provides \textbf{labeled examples} $\dataTr\smallEq\{(\bx_i,y_i)\}_i$ with $\bx \in \mathbb{R}^\din$, $y_i \in \{0,\!1\}$.
The goal of a learning algorithm is to identify a \textbf{predictor}
$f:~\mathbb{R}^\din \rightarrow \mathbb{R}$
to estimate labels%
\footnote{We define $f$ to output logits. A binary prediction $\hat{y}$ is obtained as $\hat{y}=\operatorname{round}\big(\sigma\big(f(\bx)\big)\big)$.}
of examples from a test set $\dataTest\smallEq\{\bx_i\}_i$.
While the input data $\bx$ is typically high-dimensional (\eg vectorized images),
natural data (\eg photographs) occupies only a fraction of the input space assumed to form a low-dimensional \textbf{manifold}~\cite{weinberger2006unsupervised} $\manifold \subset \mathbb{R}^\din$.
The dimensionality $\dman$ ($< \din$) is known as the \textbf{intrinsic~dimensionality} of the data.
Training and test data are drawn from a distribution on this manifold $\probId$ (in-domain examples) while unbiased natural data (free of dataset-specific selection biases) is drawn from a distribution $\probOod$ of typically broader support.

\myparagraph{Inductive biases} are the properties of a learning algorithm that determine what model $f_{\theta^\star}$ is returned for a dataset $\data$ from a hypothesis class
$\mathcal{H}=\{f_\btheta, \forall\,\btheta\}$ where $f_\btheta$ is a model with free parameters $\theta$.
Inductive biases enable generalization from finite data~\cite{mitchell1980need} by encoding assumptions on the relation between $\data$ and $\dataTest$.
In particular, classical learning theory assumes that $\data$ and $\dataTest$ contain i.i.d. samples from the same distribution. 
For completeness, we summarize a standard training workflow.

\vspace{4pt}
\begin{mdframed}[style=grayFrame]
  \small
  \begin{enumerate}[topsep=1pt,itemsep=1pt]
    \item Randomly split the data into training and validation sets: $\data \smallEq \dataTr \cup \dataVal$.
    \item A \textbf{hypothesis class} $\mathcal{H}=\{f_\btheta, \forall\,\btheta\}$ is chosen \eg by defining a neural architecture $f$.
    \item \textbf{Empirical~risk~minimization} serves to optimize the free parameters~of~$f$ as $\btheta^\star ~=~ \textrm{argmin}_\btheta \; \risk\big( f_\btheta, \color{blue}{\dataTr} \big)$
      where the empirical risk is defined as $\risk(f, \data) = \Sigma_{(\bx,y)\in\data} \; \lossPred\big(y, \sigma\big(f(\bx)\big) \big) \;/\; |\data|$,\\
      and $\lossPred$ is a predictive loss such as binary cross-entropy.
    \item \textbf{Validation performance} serves to refine various choices (architecture, regularizers, \mydots) by trial and error, \ie loosely solving $f'_{\btheta^{\star'}} ~=~ \textrm{argmin}_{f,\,\mydots} \risk(f_{\btheta^\star}, \color{blue}{\dataVal})$
      where $\risk$ is often substituted with a task-specific metric such as the error rate.
  \end{enumerate}
\end{mdframed}
There is often a multitude of models satisfying the above procedure, not all are equally desirable because they differ in properties that the procedure does not constrain.
The degree of underspecification indicates the importance of arbitrary and stochastic factors in the outcome of the learning process.

This paper focuses on \textbf{differences in OOD performance} among predictive models.
OOD~performance is the predictive performance of a model (in terms of risk, accuracy, or another task-specific metric) on test data drawn from a distribution $\probOod \smallNeq \probId$.
On OOD data, features that were predictive in the training data may become irrelevant or misleading, causing a drop in performance of a model that relies on them.
By definition, OOD performance is underspecified by the ERM objective, since the empirical risk is estimated on in-domain data.

To capture variability in OOD performance, we propose a definition of underspecification based on the number of ways to fit the data with the above procedure and produce different OOD predictions.%
\footnote{See \cite{hsu2022rashomon,semenova2019study} for related uses of the \emph{volume} of a hypothesis spaces in the definition of Rashomon sets.}

\vspace{4pt}
\begin{mdframed}[style=grayFrame]
\begin{definition}
  \label{defUnderspecification}
  The \textbf{degree of underspecification} of a dataset $\data \smallEq \dataTr \cup \dataVal$, input manifold $\manifold$, and hypothesis class $\mathcal{H}=\{f_\btheta, \forall\,\btheta\}$~
  is the ratio of volumes
  $\vol({\mathcal{H}'})/\vol(\mathcal{H})$  of the largest subset of models
  $\mathcal{H}' \subset \mathcal{H}$
  such that its elements
  $\{f_{\btheta_m}\}_m$
  all have, for small constants $\epsilon_\textrm{tr}$, $\epsilon_\textrm{val}$:
  \begin{itemize}[topsep=1pt,itemsep=1pt]
    \item A low training risk:~~ $\risk(f_{\btheta_m}, \dataTr) < \epsilon_\textrm{tr}, ~\forall \, f_\btheta \in \mathcal{H}'$,
    \item A low validation risk:~~ $\risk(f_{\btheta_m}, \dataVal) < \epsilon_\textrm{val}, ~\forall \, f_\btheta \in \mathcal{H}'$,
    \item Distinct OOD predictions:~~ $\prob
    \minSpaceSmall
    \big( \operatorname{round}(\sigma( f_{\btheta1}(\bx)
    )
    \neq
    \operatorname{round}(\sigma(
    f_{\btheta2}(\bx)
    )
    \big) \approx 1, \\
    ~\forall~f_{\btheta1}, f_{\btheta2} \in \mathcal{H}', 
    ~~f_{\btheta1}\smallNeq f_{\btheta2},
    ~~\bx \!\sim\! \probOod$.
  \end{itemize}
\end{definition}
\end{mdframed}
\vspace{-4pt}
\noindent
In the next section, we derive an algorithm to learn a set of models with these properties.

\section{Proposed method}
\label{secMethod}

\myparagraphWoSpacing{Overview.}
We train multiple models with the same architecture and data while enforcing them to represent different  functions and use different features.
The models use different initializations, but this does not always suffice to produce significantly-different models.
We add two regularizers that enforce (1)~independence of the models (mutually-orthogonal input gradients)
and (2)~alignment with the data manifold such that the models learn meaningful features.

Since the constraints directly follow from \defref{defUnderspecification}, the number of models trainable to satisfy them indicates the degree of underspecification.
The only existence of multiple such models thus highlights cases of underspecification.
The models also discover some predictive features missed by standard ERM, which can be combined by distillation into a predictor with superior OOD performance.
The next sections describe how implement the two constraints as differentiable regularizers.

\subsection{Independent models}
\label{indep}
To optimize for distinct OOD predictions, we turn the criteria of Definition~\ref{defUnderspecification} into a differentiable objective.
We reuse the concept of independent models from Ross \etal~\cite{ross2018learning,ross2020ensembles}.
\vspace{4pt}
\begin{mdframed}[style=grayFrame]
\begin{definition}
  \label{defIndep}
  A pair of predictors $f_{\btheta_1}$, $f_{\btheta_2}$ are \textbf{locally independent} at $\bx$ iff their predictions are statistically independent for Gaussian perturbations around $\bx$: 
  $f_{\btheta_1}(\widetilde{\bx})\nolinebreak\indep\nolinebreak{f}_{\btheta_2}(\widetilde{\bx})$, ~
  $\widetilde{\bx} \sim \mathcal{N}(\bx, \sigma \bI)$.
\end{definition}
\begin{definition}
  A set of predictors $\{ f_{\btheta_1}, \mydots, f_{\btheta_M}\}$ are \textbf{globally independent} on a dataset $\data$ iff every pair of them are locally independent at every $\bx \in \data$.
\end{definition}
\end{mdframed}
\vspace{-4pt}
This formalizes the notion that models can rely on different features.
In our case, we seek a set of models globally independent from one another.
We obtain a tractable objective using the relation between statistical independence and geometric orthogonality developed in~\cite{ross2018learning}.
\vspace{4pt}
\begin{mdframed}[style=grayFrame]
\begin{proposition}
  \label{mutualInformation}
  A pair of predictors $f_{\btheta_1}$, $f_{\btheta_2}$ are {locally independent} at $\bx$ (Definition~\ref{defIndep}) iff the mutual information
  $\mi(f_{\btheta_1}(\widetilde{\bx}), \,f_{\btheta_2}(\widetilde{\bx}))=0$~with $\widetilde{\bx}~\sim~\mathcal{N}(\bx, \sigma \bI)$.
\end{proposition}
\end{mdframed}
\vspace{-4pt}
\noindent
For infinitesimally small perturbations ($\sigma\minSpaceSmall\rightarrow\minSpaceSmall0$), the samples $\widetilde{\bx}$ can be approximated through linearization by the input gradients $\nabla_\bx f$.
These are 1D Gaussian random variables whose correlation is given by
$\cos\big(\nabla_\bx f_{\btheta_1}(x), \nabla_\bx f_{\btheta_2}x)\big)$
and their mutual information~\cite{gretton2003kernel}: 
$-\frac{1}{2} \ln\big(1 - \cos^2\big(\nabla_\bx f_{\btheta_1}(x), \nabla_\bx f_{\btheta_2}(x)\big)\big)$.
Therefore, the statistical independence between the models' outputs as their inputs are perturbed by small Gaussian variations can be enforced by making their input gradients orthogonal.
Our \textbf{local independence loss} for a pair of models is:
\abovedisplayskip=4pt \belowdisplayskip=4pt
\begin{align}\label{lossIndep}
  \lossIndep\big(\nabla_\bx f_{\btheta_{m_1}}\minSpaceSmall(\bx), ~ \nabla_\bx f_{\btheta_{m_2}}\minSpaceSmall(\bx)\big)
  ~=~
  \cos^2\minSpaceSmall\big(\nabla_\bx f_{\btheta_{m_1}}\minSpaceSmall(\bx), ~ \nabla_\bx f_{\btheta_{m_2}}\minSpaceSmall(\bx) \big)
\end{align}
with $\cos^2(\bv,\bw) \smallEq (\bv^\intercal \bw)^2 \, / \, (\bv^\intercal \bv)(\bw^\intercal \bw)$.
To enforce \emph{global} independence, this loss will be applied to all training points and pairs of models in the overall objective of Eq.~\eqref{lossOverall}.

\subsection{On-manifold constraint}
\label{onManifold}

\begin{wrapfigure}{R}{0.4\linewidth}
  \vspace{1.0em}
  \includegraphics[width=.95\linewidth]{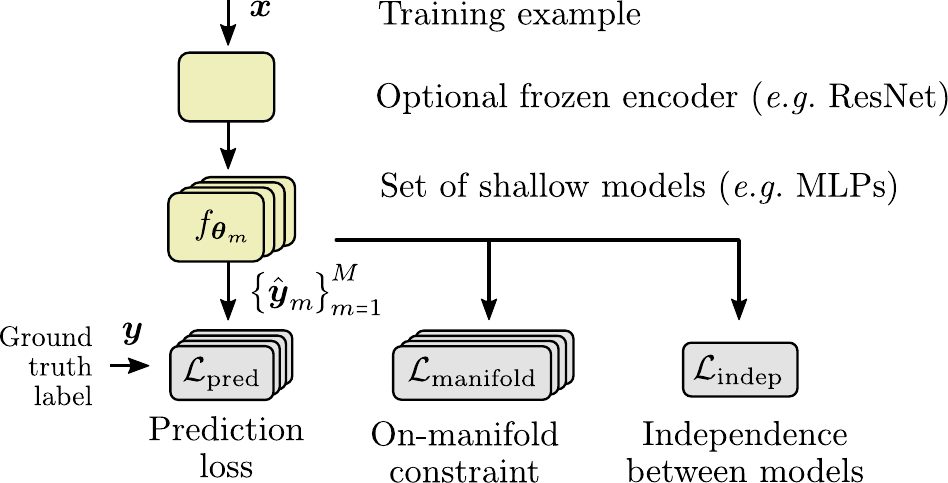}
  \caption{Method overview\\during training.}
  \vspace{0.5em}
\end{wrapfigure}
The independence constraint \eqref{lossIndep} makes models' input gradients orthogonal to one another.
The number of models satisfying it grows exponentially with the input dimension ($\din$) but many  are practically irrelevant because the natural data manifold usually occupies much fewer dimensions.
Intuitively, when the constraint affects a model's gradients in dimensions pointing outward the manifold, it does not affect its predictions on natural data.
Consequently, the independence constraint could be satisfied by models that produce identical predictions on every natural input (thus defeating its purpose) because their decision boundaries are identical when projected on the manifold.
The issue stems from the \emph{isotropic} perturbations in Eq.~\eqref{mutualInformation}. Only perturbations \emph{on} the manifold are meaningful.

One straightforward solution would be to enforce independence after projecting the data on a learned approximation of the manifold.
This approach, proposed in~\cite{ross2020ensembles,ross2018learning} failed in our early experiments because of the difficulty of optimizing the independence objective under such a strict on-manifold constraint.
Instead, we implement a soft constraint as a regularizer that proved easy to train and resilient to imperfect models of the manifold.

\begin{figure}[t!]
	\centering
  \begin{overpic}[width=0.32\linewidth]{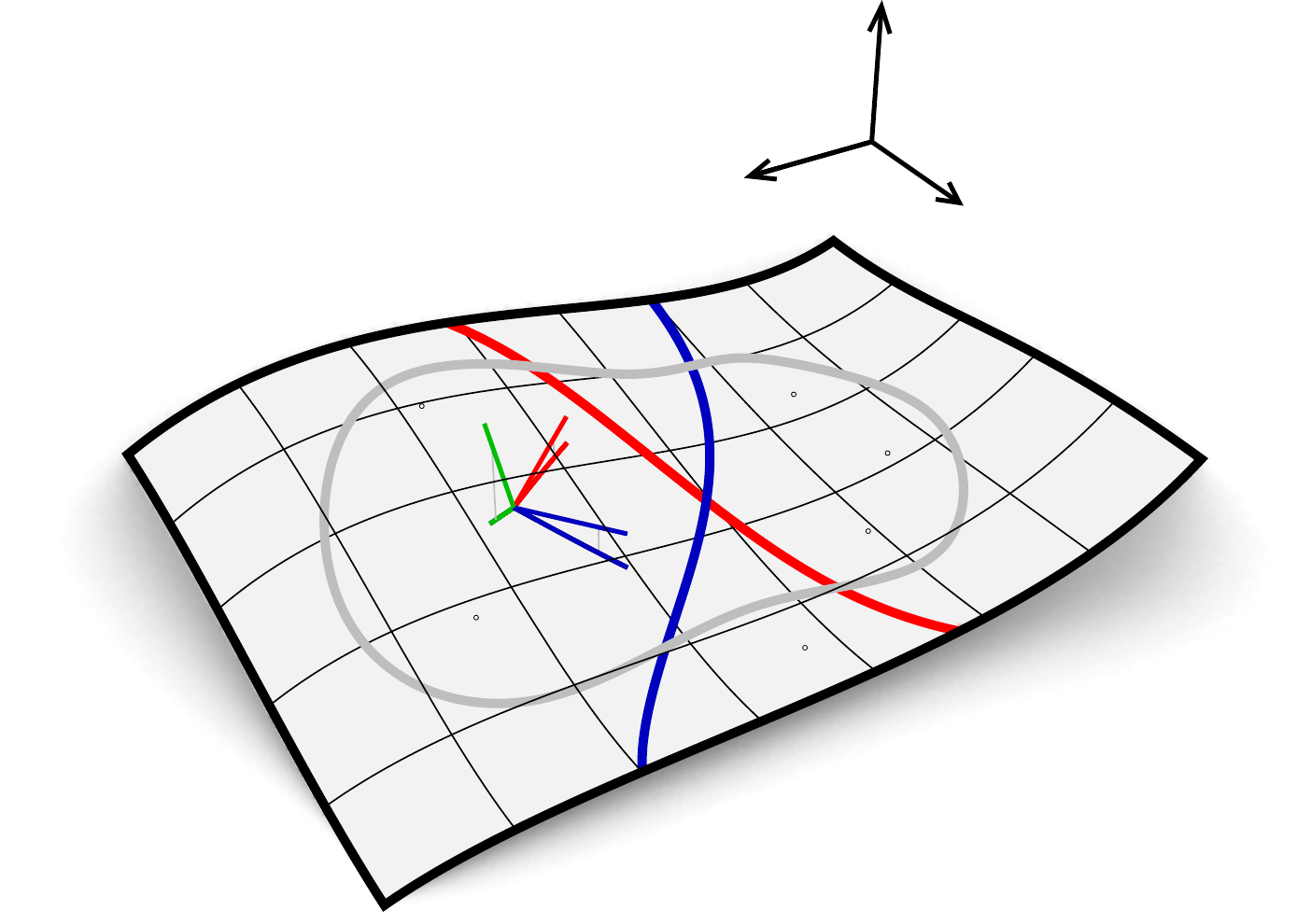}
    \put (60,39.6) {\fontsize{3.5pt}{5pt}$\blacksquare$}
    \put (68,35) {\fontsize{3.5pt}{5pt}$\blacksquare$}
    \put (66,29.6) {\fontsize{3.5pt}{5pt}$\blacksquare$}
    \put (36.0,22.6) {\fontsize{2.0pt}{5pt}$\bullet$}
    \put (31.5,39.2) {\fontsize{2.0pt}{5pt}$\bullet$}
    \put (38.8,31.35) {\fontsize{2.0pt}{5pt}$\bullet$} 
    \put (60.5,20.5) {\fontsize{6pt}{5pt}$\bigvarstar$}
    \put (74,64) {\footnotesize Input space $\mathbb{R}^\din$}
    \put (85,46) {\footnotesize Data manifold $\manifold$}
    \put (-94,64) {\scriptsize Two possible decision boundaries (\textcolor{red}{red}/\textcolor{darkBlue}{blue}),}
    \put (-94,58.65) {\scriptsize input gradients at one training point,}
    \put (-94,53) {\scriptsize and their projection on the manifold.}
    \put (56,4.7) {{\scriptsize $\bullet$} {\tiny $\blacksquare$}\scriptsize ~~Training points of two classes.}
    \put (55,-1) {\scriptsize ~~\,$\bigvarstar$~~\, Test point, OOD, underspecified.}
  \end{overpic}
	\caption{
  Effect of the proposed method in input space.
  Data such as natural images is assumed to lie on a low-dimensional manifold.
  The training set covers a subset of this manifold (gray ellipse). OOD test data ($\bigvarstar$) lies outside this subset.
  In this example, our method discovers two models (\textcolor{red}{red} and \textcolor{darkBlue}{blue} decision boundaries) whose input gradients are orthogonal (shown at one training point, in colors matching the boundary).
  Even though a third model (\textcolor{darkGreen}{green} vector) could satisfy the orthogonality constraint, its input gradient would point outside the manifold.
  This would violate the \emph{on-manifold} constraint, which requires gradients to closely match their projection on the manifold.
  \label{figManifold}\vspace{-10pt}}
\end{figure}
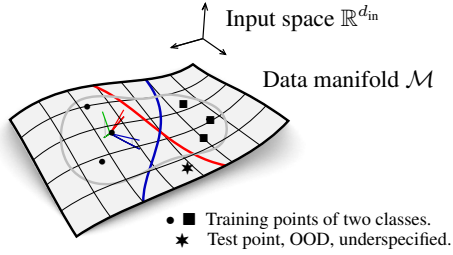

To learn the data manifold $\manifold$, we need unlabeled examples, ideally containing the type of OOD data expected at test time \eg a broad collection of natural images: $\dataUnlabelled = \{\bx_i\} \sim \probOod$.
We use this data off-line to prepare a function $\project_\manifold(\bx, \bv)$ that projects an arbitrary vector $\bv$ at $\bx$ in the input space onto the manifold (\figref{figManifold}).
During training, we penalize each model with the distance between its input gradients and their projection on the manifold.
The~\textbf{on-manifold~loss} is defined as
\begin{align}\label{lossOnManifold}
  \minSpace \lossOnManifold\big(\nabla f(\bx) \big) \smallEq
  \big|\big| \project_\manifold \big( \bx, \nabla_\bx f(\bx) \big) - \nabla_\bx f(\bx) \big|\big|_2^2.
\end{align}
\noindent
We describe several implementations of $\project_\manifold(\cdot)$ in \appref{appendixProjection} using a variational auto-encoder (VAE) or a simple PCA.
In summary, the on-manifold loss encourages a model to be sensitive to variations in the input that are likely to be encountered in natural test data.
It typically has no effect on in-domain performance (\figref{figOnManifoldAlone}) since it only removes a model's sensitivity to ``unnatural'' inputs such as variations of isolated pixels, which are unlikely to appear in natural images.

The overall learning objective combines the predictive, independence, and on-manifold losses:
\begin{align}\label{lossOverall}
  \mathcal{L}(\dataTr, \btheta_{1} \,\mydots\, \btheta_{M}) = {\Sigma}_{\bx\in\dataTr}  & \Big[~ 
  ~ ({\scriptstyle 1}/{\scriptstyle M}) \,~~{\Sigma}_{m=1}^M 
      \lossPred\big(y, \,\sigma(f_{\btheta_m}\minSpaceSmall(\bx))\big) \nonumber\\
  &+ ({\scriptstyle 1}/{\scriptstyle M^2}) ~{\Sigma}_{m_1=1}^M {\Sigma}_{m_2=1}^M 
      \wtIndep ~\, \lossIndep\big(\nabla_\bx f_{\btheta_{m_1}}\minSpaceSmall(\bx), \nabla_\bx f_{\btheta_{m_2}}\minSpaceSmall(\bx)\big) \nonumber\\
  &+ ({\scriptstyle 1}/{\scriptstyle M}) \,~~{\Sigma}_{m=1}^M 
      \wtManifold ~\, \lossOnManifold\big(\nabla_\bx f_{\btheta_m}\minSpaceSmall(\bx)\big) ~\Big].
\end{align}

\subsection{Fine-tuning}
\label{secFineTuning}

After training a set of models with \eqref{lossOverall}, we propose to relax the independence and on-manifold constraints ($\wtIndep\!\leftarrow\!0$, $\wtManifold\!\leftarrow\!0$) then fine-tune the models.
This eases the optimization and typically allows the models to reach a higher predictive accuracy.
Concretely, we apply binary masks on the data such that each model is fine-tuned only on the elements most relevant to itself:%
\footnote{In our implementation, masked elements are not replaced with zeros, but rater with random values from other instances in the current mini-batch.}
\abovedisplayskip=5pt \belowdisplayskip=4pt
\begin{align}
  \dataTr^m = \{(\bx_i \odot \mask^m_i, y_i) : (\bx_i, y_i) \in \dataTr\}
\end{align}
with $\mask^m_i \in \{0,1\}^\din$.
The masks are computed before starting the fine-tuning to highlight the data most relevant to each model.
Each element (i.e. pixel or channel) is unmasked only for the model with the largest corresponding gradient magnitude:
\abovedisplayskip=5pt \belowdisplayskip=3pt
\begin{align}\label{eqMasks}
  \minSpaceSmall\minSpace\mask^m_i = \indicator\big( m \smallEq \argmax_{1 \leq m \leq M} \nabla f_{\btheta_m}\minSpaceSmall(\bx_i) \big) ~~\,\forall \, (\bx_i, \cdot) \in \dataTr.
\end{align}
We fine-tune each model on its own masked version of the data.%
\footnote{We obtain very similar results between fine-tuning and retraining models from scratch on the masked data.}
This ensures that the models remain distinct despite disabling the regularizers ($\wtIndep\!\leftarrow\!0$, $\wtManifold\!\leftarrow\!0$).
See \algref{algTraining} for a summary.

\begin{algorithm}[t!]
  \footnotesize
  \caption{Training and fine-tuning models.}
  \label{algTraining}
  \DontPrintSemicolon
    \textbf{Inputs:} Labeled examples $\dataTr$.
    Unlabeled examples $\dataUnlabelled$ (typically $\dataTr\subset\dataUnlabelled$). Architecture $f$.\\
    \textbf{Result:}
    Set of independent models $\{f_{\btheta_1} \, \mydots \, f_{\btheta_M}\}$.\\
  \textbf{Method:}\\
  With $\dataUnlabelled$, estimate dimensionality $\dman$~\cite{pope2021intrinsic}~~and set the number of models $M \leftarrow \dman$.\\
  With $\dataUnlabelled$, prepare function $\project(\cdot)$~~by PCA decomp. or by training a VAE.\\ 
  With $\dataTr$, train $M$ instances of $f$ in parallel (Eq. \ref{lossOverall}): 
  $\{\btheta_1 \, \mydots \, \btheta_M\}  \leftarrow \argmin \mathcal{L}(\dataTr, \btheta_{1} \, \mydots \, \btheta_{M}).$\\
  Determine masks on input data (Eq. \ref{eqMasks}): $\{ \mask^m_i \}_{i,m}$\\
  \ForEach(~~\myComment{Optional fine-tuning on masked data}){$m$}{
    $\dataTr^m \leftarrow \{(\bx_i \odot \mask^m_i, y_i)\}_i$~~\myComment{Prepare masked data}\\
    $\wtIndep \leftarrow 0, ~\wtManifold \leftarrow 0$~~\myComment{Use only predictive loss}\\
    $\btheta_m \leftarrow \argmin \mathcal{L}(\dataTr^m, \btheta_{m})$~~\myComment{Fine-tune}\\
  }
  \BlankLine
\end{algorithm}

\subsection{Distilling multiple models into one}
\label{secCombination}
Finally, after training/fine-tuning a set of models, we propose to combine the best of them into a global one that uses all of the most relevant features.%
\footnote{Our combination of models is not equivalent to an ensemble: we train a global model to use/ignore specific, chosen features. In comparison, a traditional ensemble combines models in prediction space, relying on their uncorrelated errors to lower the prediction variance. See also \appref{disentanglement} for an the analogy of our method with disentanglement in predictor space.}
We train this global model from scratch, without regularizers, on masked data as described above, using masks from \emph{multiple} selected models combined with a logical OR.
In our experiments, we combine the two models with the highest accuracies on an OOD validation set.
We repeat this pairwise combination as long the accuracy of the global model increases, usually for 2--3 iterations (as formalized in \algref{algCombining} in the Appendix).



\section{Practical considerations}

\myparagraphWoSpacing{Number of models to train.}
\label{secQuantification}
We initially estimate the dimensionality of the manifold ($\dman$) from $\dataUnlabelled$.
This is the upper bound on the number of predictors aligned with the manifold that can be mutually independent.
\label{secIntrinsicDimensionalityEstimation}%
The dimensionality is estimated with a simple method~\cite{pope2021intrinsic} based on distances between nearest neighbours.
Only the worst case (highest degree of underspecification per \defref{defUnderspecification}) should allow training a number of models $M\smallEq\dman$ while satisfying the independence and on-manifold constraints.
In our experiments, underspecification is usually less severe.
We observed that the value of the predictive loss ($\lossPred$) on training and validation data would remain high for some models, while it would converge normally for others.
Therefore, monitoring the loss per model could potentially serve for evaluating the degree of underspecification of a given setting (dataset and architecture) by noting the number of models that converge.
A thorough investigation and validation of this technique is an important topic left for future work.

\myparagraph{Is the data manifold necessary in the definition of underspecification~?}
Our definition without the reference to the data manifold would be pointless because virtually all conceivable models would be severely underspecified.
The manifold (\eg natural photographs) only fills a fraction of the ambient input space (\eg all possible RGB pixel grids).
Our definition captures underspecification with respect to this smaller range of inputs that are expected at test time.

\myparagraph{How to select models to combine into a global predictor?}
\label{secSelectingFeatures}
Our approach is agnostic to the model selection strategy.
Many options are possible to identify models that rely on ``robust'' features.
The simplest is an evaluation on an OOD validation set.
Alternatively, interpretability methods allow domain experts to inspect models and the features they rely on.
The selection may also be semi-automated with task-specific heuristics~\cite{deng2021does,garg2021ratt,immer2021scalable,wald2021calibration} or additional annotations \eg human attention and rationales~\cite{selvaraju2019taking,stacey2021natural}.


\myparagraph{Computational cost.}
With parallelization, our method scales sublinearly in computing time and linearly in memory w.r.t. the number of models.
Models share mini-batches.
And since they use the same architecture, most computations can be parallelized with \textbf{grouped convolutions} (one group per model).
The method is also applicable on a frozen feature extractor (\secref{secCamelyon} and \appref{appendixCollagesResNet}).
However, fine-tuning a shared extractor with the independence loss is not possible. The model could simply dispatch redundant features on multiple channels and satisfy the independence constraint without the desired increase in diversity.
Compared to~\cite{teney2021evading}, we significantly reduce the number of models needed to discover useful features.

\section{Experiments}
\label{secExp}

We first present experiments that validate the method on controlled data with multiple known predictive features (collages, \secref{secCollages}).
We then demonstrate applications to existing datasets: WILDS-Camelyon17, \ref{secCamelyon} and GQA, \secref{secVqa}.

\subsection{Experiments on controlled data: collages}
\label{secCollages}

This diagnostic dataset contains images with binary labels that are constructed to contain multiple predictive features~\cite{shah2020pitfalls,teney2021evading}.
Each image contains four tiles representing one of two classes respectively from MNIST (0/1), CIFAR-10 (automobile/truck), Fashion-MNIST (pullover/coat), and SVHN (0/1).
\begin{itemize}[topsep=3pt,itemsep=3pt]
  \item At \textbf{training time}, the labels are perfectly correlated with the four tiles (0/1 respectively for the first/second possible class in each tile).
  There are (at least) four equally-valid ways of understanding the task (\ie relying on any of the four tiles).
  \item At \textbf{test time}, we evaluate a model on four test sets that represent different OOD conditions. In each, only one tile is correlated with the correct label while others tiles are randomized.
  By examining the performance on the four test sets, we can identify which tile(s) the model relies on
\end{itemize}
\floatsetup[figure]{capposition=beside,capbesideposition={center,right},floatwidth=0.54\textwidth,capbesidewidth=0.42\textwidth}
\begin{figure}[h!]
  \centering
  \renewcommand{\tabcolsep}{.1em}
  \renewcommand{\arraystretch}{1}
  \begin{tabular}{rcc rcc}
    {\begin{tabular}{r}\scriptsize \textbf{Class 0}\vspace{-4pt}\\\scriptsize Zero, pullover\vspace{-5pt}\\\scriptsize automobile, zero. \end{tabular}} & \includegraphics[align=c,width=.14\linewidth]{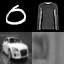} & \includegraphics[align=c,width=.14\linewidth]{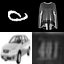} &
    {\begin{tabular}{r}\scriptsize \textbf{Class 1}\vspace{-4pt}\\\scriptsize One, coat\vspace{-5pt}\\\scriptsize truck, one.            \end{tabular}} & \includegraphics[align=c,width=.14\linewidth]{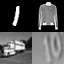} & \includegraphics[align=c,width=.14\linewidth]{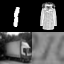}\\
  \end{tabular}
  \vspace{-12pt}
  \caption{Examples of collages~\cite{teney2021evading}. Training labels are correlated with all four tiles. Each tile contains features of different complexity.\vspace{-15pt}}
  \label{figCollagesExamples}
\end{figure}
\vspace{7pt}


\myparagraph{Task difficulty.}
This dataset is surprisingly challenging because the tiles vary greatly in learning difficulty (\eg MNIST 0s/1s are very distinct while Fashion-MNIST pullovers/coats look extremely similar).
It would be reasonable to learn a model that relies on all four tiles. However, an ERM-trained baseline surprisingly uses only a few MNIST pixels (achieving $\sim$99\% accuracy on the MNIST test set and $\sim$50\% on the others), as shown in previous work on the simplicity~bias of neural networks~\cite{teney2021evading}.

We follow~\cite{teney2021evading} and use our method to learn multiple models compatible with the data. We then report the accuracy of the best model on each test set, \ie the best accuracy assuming perfect model selection.
This avoids confounding the performance of the learning algorithm and with that of the selection strategy.

\myparagraph{Applying the proposed method.}
We follow \algref{algTraining}.
We prepare unlabeled data to defines the data manifold as the union of the training and test sets, thus covering all combinations of contents of the four tiles.
With this data, we estimate the dimensionality of the manifold with~\cite{pope2021intrinsic} as about 23.8 ($\sigma\smallEq0.16$ over 10 runs).
We prepare two generative models of the manifold: a PCA with 24 components (capturing $\sim$85\% of the variance)
and a VAE with 24 latent dimensions (details in \appref{appendixImplementation}).
We define a simple architecture (2-layer MLPs) and train multiple instances in parallel with the proposed objective.
The only hyperparameters are the number of models and weights of independence/on-manifold constraints.
We plot a range of values in the appendix (\figref{figCollagesHeatmaps2}).

\begin{figure*}[t!]
  \vspace{-8pt}
  \centering
  \renewcommand{\tabcolsep}{0em}
  \renewcommand{\arraystretch}{1.1}
  \tiny
  \begin{subfigure}[t]{0.29\linewidth}
    \centering
    \begin{tabular}{rc}
      & \textbf{Baseline}\vspace{2pt}\\
      { 4 Models~~} & \hangBoxC{\includegraphics[width=70pt]{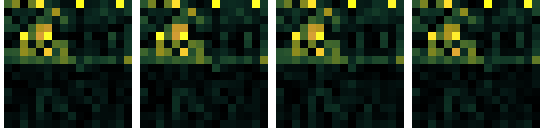}}\\
      {8 Models~~} & \hangBoxC{\includegraphics[width=70pt]{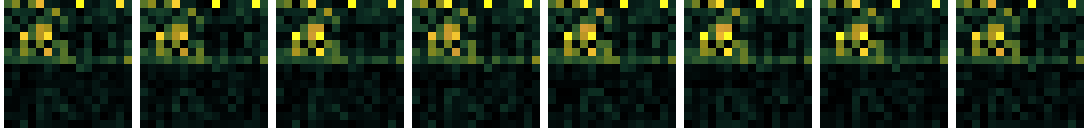}}\\
      {12 Models~~} & \hangBoxC{\includegraphics[width=70pt]{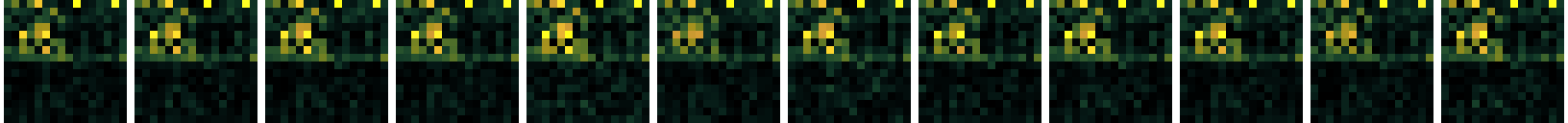}}\\
    \end{tabular} 
    \caption{\scriptsize With standard training, all models rely on a small, identical region of the image, despite the fact that predictive features are present all over.}
  \end{subfigure}\hspace{5pt}
  \begin{subfigure}[t]{0.21\linewidth}
    \centering
    \begin{tabular}{c}
      \textbf{+ Independence}\vspace{2pt}\\
      \hangBoxC{\includegraphics[width=70pt]{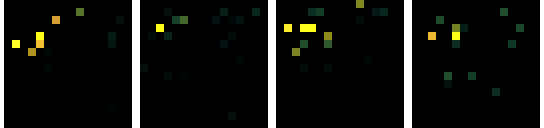}}\\
      \hangBoxC{\includegraphics[width=70pt]{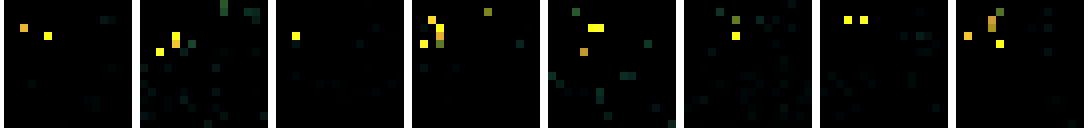}}\\
      \hangBoxC{\includegraphics[width=70pt]{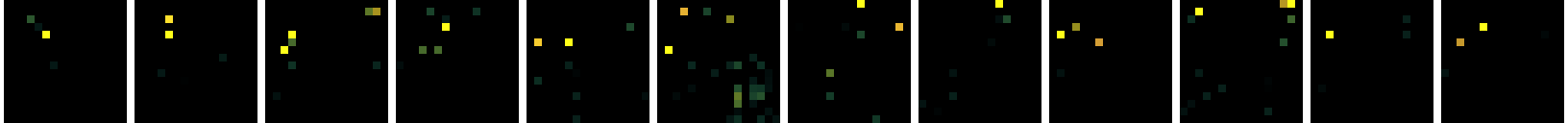}}\\
    \end{tabular} 
    \caption{\scriptsize Independence produces distinct gradients, but many models are needed to discover new features and they are sensitive to isolated pixels.}
  \end{subfigure}\hspace{5pt}
  \begin{subfigure}[t]{0.21\linewidth}
    \centering
    \begin{tabular}{c}
      \textbf{+ On-manifold}\vspace{2pt}\\
      \hangBoxC{\includegraphics[width=70pt]{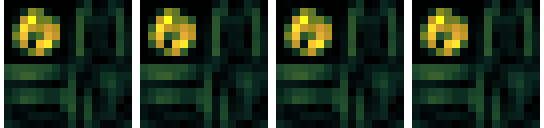}}\\
      \hangBoxC{\includegraphics[width=70pt]{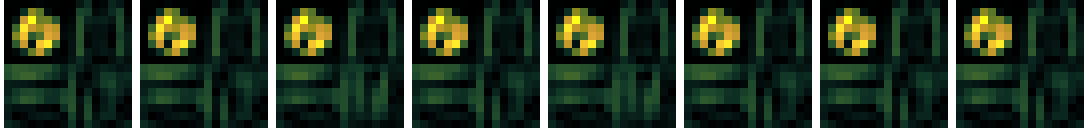}}\\
      \hangBoxC{\includegraphics[width=70pt]{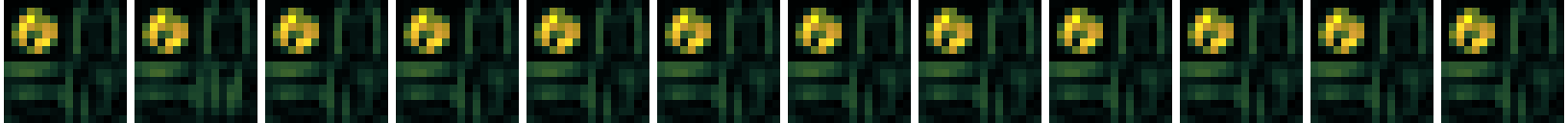}}\\
    \end{tabular} 
    \caption{\scriptsize The on-manifold constraint forces gradients to align with natural variations of the data. Accuracy is virtually identical to the baseline.\label{figOnManifoldAlone}}
  \end{subfigure}\hspace{5pt}
  \begin{subfigure}[t]{0.21\linewidth}
    \centering
    \begin{tabular}{c}
      \textbf{+ Indep. + on-manifold}\vspace{2pt}\\
      \hangBoxC{\includegraphics[width=70pt]{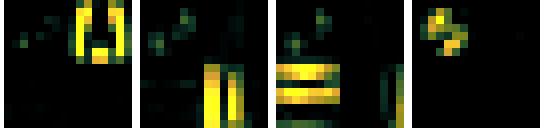}}\\
      \hangBoxC{\includegraphics[width=70pt]{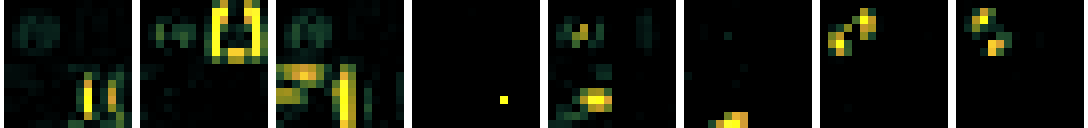}}\\
      \hangBoxC{\includegraphics[width=70pt]{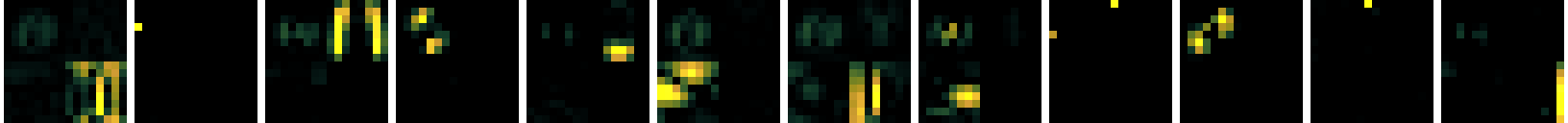}}\\
    \end{tabular} 
    \caption{\scriptsize With both constraints, \textbf{we learn semantically relevant features in all image regions} with as few as 4 models.}
  \end{subfigure}
  \vspace{-5pt}
  \normalsize
  \caption{Input gradients for a random test image from the collages dataset.
  It is remarkable that, with the proposed method~(d) \textbf{image regions emerge as meaningful features without any inductive bias for spatial locality} such as convolutions (models in these experiments are fully-connected MLPs).
  \label{figCollagesGradients}\vspace{-10pt}}
\end{figure*}

\myparagraph{Results.}
Our method learns a set of models that focus on different parts of the images. 
Remarkably, learning as few as 4 models is sufficient to obtain models with high accuracy on all of the four test sets.
Let us examine several ablations.
\begin{itemize}[topsep=3pt,itemsep=3pt]
  \item The baseline ($\wtIndep\smallEq\wtManifold\smallEq0$) only learns about MNIST. 
  \item The independence constraint ($\wtIndep\minSpaceSmall>\minSpaceSmall$, $\wtManifold\smallEq0$) is crucial for learning distinct models. On its own, it requires training a very large number of models ($\gg$32) before picking up features outside the MNIST tiles.
  Visualizations of input gradients (\figref{figCollagesGradients}) reveal that these models each rely only on a single or a few pixels.
  These trivial solutions to the independence constraint, akin to adversarial examples, are avoided with the on-manifold constraint.
  \item In the full method ($\wtIndep\minSpaceSmall>\minSpaceSmall0$, $\wtManifold\minSpaceSmall>\minSpaceSmall0$) the models discover distinct features that align with the semantic contents of images. 
  The effect of the on-manifold constraint on input gradients is striking (\figref{figCollagesGradients}).
  It forces models to be sensitive to natural variations of the data --~rather than unlikely single-pixel patterns.
  Remarkably, \textbf{image regions emerge as meaningful features without inductive bias for spatial locality} (\eg no convolutions). 
\end{itemize}
\floatsetup[figure]{capposition=beside,capbesideposition={center,right},floatwidth=0.35\textwidth,capbesidewidth=0.61\textwidth}
\begin{figure}[h!]
  \includegraphics[width=\linewidth]{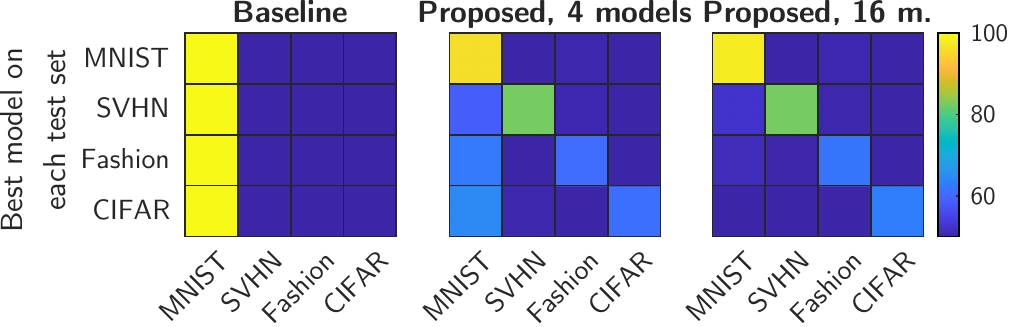}
  \vspace{-10pt}
  \caption{Collages dataset: accuracy on the four test sets~(columns) of models with best accuracy on each set~(rows).
  Diagonal patterns indicate that \textbf{models specialize and learn different, non-overlapping features}.
  The baseline only learns features relevant to MNIST.
  \label{figCollagesMatrices}}
  \vspace{-5pt}
\end{figure}

\myparagraph{Hyperparameters.}
A number of models between 4 and 24 give excellent results. As expected, the larger this number, the more granular the features these models learn (\figref{figCollagesGradients}).
The effect breaks down for $>$24 models, matching theoretical expectations since the dimensionality of the manifold was estimated at $\sim$24.
The method is stable over a range of regularizer weights.
Additional comparisons in \tabref{tabCollagesPixels} show that a VAE is better than PCA to represent the manifold.
This agrees with the general expectation that natural images form a non-linear manifold in pixel space.
We also found overall results to be robust to variations in architecture and hyperparameters of the VAE. 

\floatsetup[figure]{capposition=below,floatwidth=\textwidth}

\myparagraph{Fine-tuning.}
We report the accuracy of models fine-tuned on masked inputs as proposed in \secref{secFineTuning}.
This optional step relaxes the independence constraint to maximize each model's predictive performance.
The accuracy jumps significantly and almost reaches the upper-bound on each test set (\tabref{tabCollagesPixels}).
We experimented with relaxing both the independence and on-manifold constraints. Disabling the former has a significant effect. But the latter has no significant effect on accuracy on its own as expected and discussed in \secref{onManifold}.

\myparagraph{Distilling multiple models into one.}
We report the performance of combinations of features described in \secref{secCombination}.
This procedure is most effective after training a large number of models (24 here).
This is unsurprising since models then discover finer-grained features.
Each combination selects features relevant to only one specific tile to achieve near-maximal accuracy on the test set of that tile.
Simple traditional ensembling of models completely failed in our experiments.

\myparagraph{Comparison with existing methods.}
No other method reported in \tabref{tabCollagesPixels} performed well on this dataset.
The method of Teney \etal~\cite{teney2021evading} is technically the most similar to ours, but it requires training a much larger number of models and still achieves much lower accuracy.
While all experiments of this section used a model taking raw pixels as input, we repeated the whole evaluation using a shared, frozen ResNet to extract features in \appref{appendixCollagesResNet}.
This implementation is computationally appealing for larger-scale applications, and gave essentially similar findings with higher overall accuracy thanks to the deeper architecture.

\floatsetup[table]{capposition=beside,capbesideposition={center,right},floatwidth=0.55\textwidth,capbesidewidth=0.4\textwidth}
\begin{table}[h!]
  \vspace{-8pt}
  \caption{
    Accuracy on \textit{collages} of existing and proposed methods (8 models per method unless specified).
    The 4 test sets simulate different OOD conditions: only one tile in each set is correlated with the labels.
    \textbf{Standard training only learns a fraction of predictive features}.
    Existing methods cannot do better than chance except on MNIST, or they require training a large number of models.
    Ours learns a variety of features and give near-optimal predictions on every test set (last row).
    \label{tabCollagesPixels}}
  \tiny
  \renewcommand{\tabcolsep}{0.50em}
  \renewcommand{\arraystretch}{1.0}
  \centering
  \begin{tabularx}{\linewidth}{Xccccc}
  \toprule
  \textbf{Collages} dataset (accuracy in \%) & \multicolumn{4}{c}{Best model on} \vspace{1.5pt}\\ \cline{2-5}
  ~ &  \rotatebox{90}{MNIST} & \rotatebox{90}{SVHN~~} & \rotatebox{90}{Fashion\phantom{e}} & \rotatebox{90}{CIFAR-10\phantom{e}} & \rotatebox{90}{Average}\\
  \midrule
  Upper bound (training on test-domain data) & 99.9 & 92.4 & 80.8 & 68.6 & 85.5 \\
  \midrule
  ERM Baseline & 99.8 & 50.0 & 50.0 & 50.0 & 62.5 \\
  Spectral decoupling~\cite{pezeshki2020gradient} & 99.9 & 49.8 & 50.6 & 49.9 & 62.5 \\ 
  With penalty on L1 norm of gradients & 98.5 & 49.6 & 50.5 & 50.0 & 62.1 \\ 
  With penalty on L2 norm of gradients~\cite{hoffman2019robust} & 96.6 & 52.1 & 52.3 & 54.3 & 63.8 \\ 
  Input dropout (best ratio: 0.9) & 97.4 & 50.7 & 56.1 & 52.1 & 64.1 \\ 
  Independence loss (cosine similarity) \cite{ross2020ensembles} & 99.7 & 50.4 & 51.5 & 50.2 & 63.0 \\ 
  Independence loss (dot product) \cite{teney2021evading} & 99.5 & 53.5 & 53.3 & 50.5 & 64.2 \\ 
  \midrule
  With {many more} models\\
  Independence loss (cosine similarity), \underline{1024} models & 99.5 & 58.1 & 66.8 & 63.0 & 71.9 \\ 
  Independence loss (dot product), \underline{128} models & 98.7 & 84.9 & 71.6 & 61.5 & 79.2 \\ 
  \midrule
  Proposed method (only 8 models)\\
  Independence + on-manifold constraints, PCA & 97.3 & 69.8 & 62.2 & 60.0 & 72.3 \\ 
  Independence + on-manifold constraints, VAE ($^\ast$) & 96.5 & 85.1 & 61.1 & 62.1 & 76.2 \\ 
($^\ast$) ~+~ FT ~(fine-tuning) & 99.7 & 90.9 & 81.4 & 67.4 & 84.8 \\ 
  ($^\ast$) ~+~ FT ~+~ pairwise combinations (1$\times$) & 99.9 & 92.2 & 79.3 & 66.3 & 84.4 \\ 
  ($^\ast$) ~+~ FT ~+~ pairwise combinations (2$\times$) & 99.9 & 92.5 & 80.2 & 67.5 & 85.0 \\ 
  \textbf{($^\ast$) ~+~ FT ~+~ pairwise combinations (3$\times$)} & \textbf{99.9} & \textbf{92.3} & \textbf{80.8} & \textbf{68.5} & \textbf{85.4} \\ 
  \bottomrule
  \end{tabularx}
  \vspace{-10pt}
\end{table}

\subsection{Experiments on real data: WILDS-Camelyon17}
\label{secCamelyon}

\myparagraph{Dataset.}
The WILDS-Camelyon17 benchmark~\cite{koh2020wilds} provides histopathology images to classify as ``\textit{tumor}'' or ``\textit{normal}''.
The images come from different sets of hospitals in the training, validation (val-OOD), and test splits (test-OOD).
The challenge is to learn a model that generalizes from the training hospitals to those of the test set.
The original authors of~\cite{koh2020wilds} trained a Densenet-121 model from scratch on this data with 10 random seeds. They showed that the performance on val-OOD and test-OOD varies wildly across seeds, demonstrating that the task is severely underspecified with only the standard training images (the dataset provides additional hospital labels that could enable generalization; neither ERM nor our method uses them).


\myparagraph{Implementation of our method.}
We use frozen features (last-layer activations) from one of the pretrained models from~\cite{koh2020wilds} as input. We will show that we can recover even more variability in performance than the complete models trained on different random seeds, even while keeping the model frozen (\ie retraining only a classifier).
We first determined that the best ERM-trained classifier on frozen features is a simple linear one, rather than an MLP.
Our method simplifies in two ways with a linear classifier.
First, input gradients are equal to the classifier weights, and the proposed regularizers do not require second-order derivatives anymore during back-propagation.
Second, we found empirically that the soft on-manifold regularizer can be replaced with a hard constraint: we explicit project the input gradients onto the manifold and apply the independence regularizer on these projections, as proposed in~\cite{ross2020ensembles}.
As noted in \secref{onManifold}, this option completely failed in our early experiments with MLPs, but it seems viable with linear classifiers. This further simplifies the implementation and removes the hyperparameter $\wtManifold$.

\begin{figure}[p]
  \centering
  \includegraphics[width=1\textwidth]{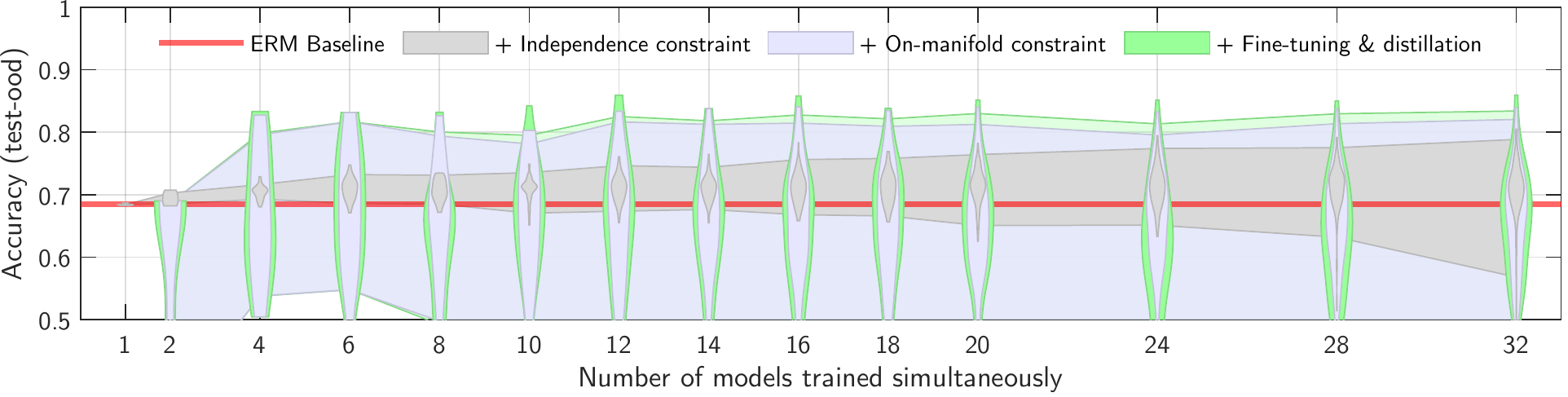}
  \vspace{-5pt}
  \caption{
  Spread of accuracies on WILDS-Camelyon17 of models trained with different ablations of our method.
  The upper/lower bounds of the \textbf{shaded~areas} show the highest/lowest accuracy of any model from one run, averaged over 6 seeds.
  The \textbf{violins} show the distribution of accuracies over {all} seeds (hence some values outside the shaded areas).
  \textbf{Take-aways.}~The independence constraint (\textbf{gray}) produces a wide variety of models compared to the baseline (\textbf{red}).
  However, the highest accuracy in each run grows slowly with the number of models.
  {With the on-manifold constraint}~(\textbf{blue}), the improvement is clearly larger and only requires a handful of models. The fine-tuning and distillation~(\textbf{green}) bring an additional marginal improvement.
  \label{figCamelyon}\vspace{0pt}}
\end{figure}

\begin{figure}[p]
  \centering
  \includegraphics[height=130pt]{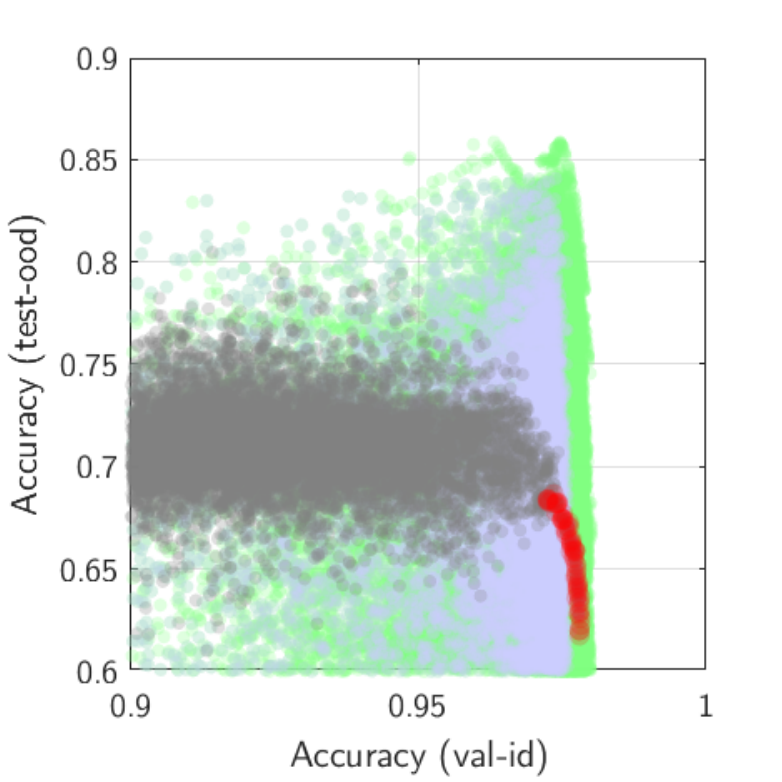}~
  \includegraphics[height=130pt]{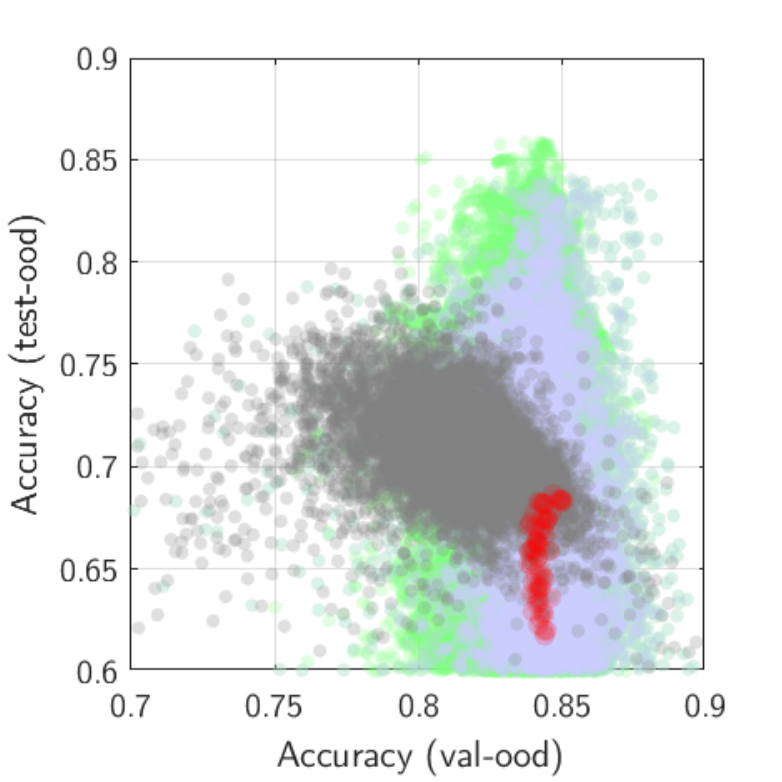}
  \vspace{-5pt}
  \caption{On WILDS-Camelyon17, each model (same colors as \figref{figCamelyon}) achieves a different trade-off in performance on different splits, as seen with test-OOD \textit{vs.} val-ID/val-OOD (left/right, respectively).
  Each dot represents one model at one training epoch. We plot every epoch since each model moves on these charts as training progresses.
  The proposed method (green) produces models with much better performance on all splits (toward the upper right) than the baseline (red).
  \label{figCamelyonTradeoff}\vspace{0pt}}
\end{figure}

\myparagraph{Results.}
We plot in \figref{figCamelyon} the spread of accuracies of models trained with different methods (using features from the first pretrained model from~\cite{koh2020wilds}, see \appref{appendixWilds} for similar results with the others).
The \textbf{ERM baseline} simply recovers the accuracy of the original complete Densenet, with essentially no variation across random seeds.
With our \textbf{independence constraint}, the spread of accuracies significantly widens, both below and above the baseline.
In \figref{figCamelyonTradeoff}, we see that the models span various trade-offs in accuracy on val-OOD and on test-OOD, neither of which is correlated with the accuracy on in-domain data (val-ID), thus showing evidence of underspecification.
Back to \figref{figCamelyon}, with our additional \textbf{on-manifold constraint}, the best models reach higher accuracies. This also tops out when training a handful of models (about 10--14, near the intrinsic dimensionality of the data estimated at 12 with~\cite{pope2021intrinsic}).
Keeping in mind that we use frozen features, these results show that the ERM-pretrained model extracts features useful for OOD performance but that are ignored by the pretrained classifier.
Similar findings were recently reported in~\cite{kirichenko2022last,rosenfeld2022domain}.
Our method recovers these features and produces alternative classifiers with a variety of trade-offs in performance across various OOD conditions.

\myparagraph{Ablations.}
In \tabref{tabCamelyon}, we compare additional ablations of our method, using a fixed number of 10 models.
The \textbf{essential components are the independence and on-manifold constraints}.
The fine-tuning and distillation steps contribute to a marginal improvement. 
We report similar \emph{relative} improvements in \appref{appendixWilds} with other pretrained models, but the absolute performance is very much \textbf{dependent on a ``good'' pretrained model}.

\myparagraph{Model selection.}
Our method brings similar relative improvements on either val-OOD or test-OOD but typically with different models for each (see \figref{figCamelyon}) despite both being OOD relative to the training data.
\textbf{Model selection absent labelled target-domain data therefore remains an issue} on this dataset.
Fortunately, only little such data may be sufficient.
We repeated a few experiments while holding out 1\% of test-OOD (less than 1,000 instances) and we observed a 99.87\% correlation coefficient between the accuracy on test-OOD and this held-out data.
While all our results assume perfect ``oracle'' model selection, it seems reasonable that real applications could provide a small amount of labelled test data to achieve similar results.

\floatsetup[table]{capposition=beside,capbesideposition={center,right},floatwidth=0.58\textwidth,capbesidewidth=0.40\textwidth}
\begin{table}[p]
  \caption{Accuracy on WILDS-Camelyon17 (average over 6 random seeds) while training 12 models with various ablations.
  Each component of our method contributes to improving the accuracy of the best model from each run. The data also appears simple enough that a PCA approximates the manifold well enough. This allows implementing the on-manifold constraint as a hard projection rather than a soft regularizer.\label{tabCamelyon}}
  \small
  \renewcommand{\tabcolsep}{0.15em}
  \renewcommand{\arraystretch}{1.05}
  \centering
  \begin{tabularx}{\linewidth}{Xccc}
    \toprule
    \textbf{WILDS-Camelyon17} & \multicolumn{3}{c}{Best accuracy (\%) on}\\
    ~ & val-OOD & ~ & test-OOD\\
    \midrule
    Pseudo-Label~\cite{lee2013pseudo} & -- && 67.7 \pp{8.2} \\
    DANN~\cite{ganin2016domain} & -- && 68.4 \pp{9.2} \\
    FixMatch~\cite{sohn2020fixmatch} & -- && 71.0 \pp{4.9} \\
    CORAL~\cite{sun2017correlation} & -- && 77.9 \pp{6.6} \\
    NoisyStudent~\cite{xie2020self} & -- && \underline{\textbf{86.7}} \pp{1.7} \\
    \midrule
    ERM Baseline                               & 84.9 \pp{0.1} && 68.4 \pp{0.1} \\
    + Independence constraint                  & 85.3 \pp{0.5} && 74.6 \pp{0.9} \\
    + On-manifold soft regularizer, VAE        & 85.4 \pp{0.4} && 80.3 \pp{1.7} \\
    + On-manifold hard projection, VAE         & 88.2 \pp{2.1} && 76.3 \pp{2.8} \\
    + On-manifold soft regularizer, PCA        & 87.8 \pp{0.3} && 79.0 \pp{2.9} \\
    + On-manifold hard projection, PCA$^\ast$ & \textbf{88.4} \pp{0.7} && 81.6 \pp{1.4} \\
    ($^\ast$) + Fine-tuning \& distillation      & \textbf{88.4} \pp{0.7} && \textbf{82.5} \pp{2.4} \\
    \bottomrule
  \end{tabularx}
  \vspace{-10pt}
\end{table}

\clearpage
\subsection{Experiments on visual question answering: GQA dataset}
\label{secVqa}

We now demonstrate the applicability of our method on a more complex task.
We choose visual question answering (VQA) because it is notorious for dataset biases~\cite{teney2020value} that cause  shortcut learning~\cite{dancette2021beyond}.
We train a simple architecture (details in \appref{appendixVqa}) on the GQA dataset~\cite{hudson2018gqa} and report accuracy on the GQA and GQA-OOD~\cite{kervadec2021roses} validation sets.
We focus on binary questions to lighten the computational expense.
As input the model, we use frozen 2048-dimensional image features (globally-pooled bottom-up features~\cite{anderson2017features}).
Our method is applied in the space of these features.
In \tabref{tabVqa}, we observe that our method produces models with slightly better accuracy on all test sets.
The independence regularizer alone is not sufficient, most likely because it is unable to discover meaningful features in the large 2048-dimensional space.
The on-manifold constraint does solve this difficulty.
Interestingly, the individual models are better as well as a simple ensemble of all of these models.
This suggests a benefit from the greater variety of features learned collectively by the models.
We show in \figref{figTeaser} and in \appref{appendixVqaResults} that the models typically focus each on different regions of images (with grad-CAM-weighted visualizations as in~\cite{selvaraju2019taking}).
It is important to note that our method is applied across channels of image features, and that the spatial diversity emerges naturally.
We quantitatively verify this increase in diversity in \tabref{tabVqa} with the average Spearman rank correlation of grad-CAM scores across models. 
\floatsetup[table]{capposition=beside,capbesideposition={center,right},floatwidth=0.61\textwidth,capbesidewidth=0.36\textwidth}
\begin{table}[h!]
  \vspace{-8pt}
  \caption{Application to visual question answering. Models trained with the proposed method achieve higher accuracy.
  They also show higher diversity in the image regions they rely on (last column, lower correlation of grad-CAM scores).\label{tabVqa}}
  \vspace{4pt}
  \tiny
  \renewcommand{\tabcolsep}{0.15em}
  \renewcommand{\arraystretch}{1.0}
  \centering
  \begin{tabularx}{\linewidth}{Xcccccc}
  \toprule
  \textbf{GQA yes/no} & N. of & GQA Val. & GQA Val. & GQA-OOD & GQA-OOD & Grad-CAM\\
  (accuracy in \%) & models & \scriptsize (best) & \scriptsize(ensemble) & Val-head \scriptsize (best) & Val-tail \scriptsize (best) & rank corr.\\
  \midrule
  Baseline &  3 & 67.9 & 69.3 & 70.4 & 66.5 & 0.68 \\
  Baseline & 16 & 68.6 & 68.9 & 71.2 & 67.3 & -- \\
  \midrule
  + Independence &  3 & 67.6 & 69.2 & 70.5 & 67.1 & 0.57 \\
  + Independence &  4 & 67.6 & 69.4 & 70.1 & 66.2 & -- \\
  + Independence &  8 & 67.8 & 70.0 & 70.7 & 66.7 & -- \\
  + Independence & 12 & 68.1 & 70.3 & 70.6 & 69.8 & -- \\
  + Independence & 16 & 68.1 & \textbf{70.5} & 71.1 & 69.6 & -- \\
  \midrule
  + Ind. + on-manifold &  3 & 68.7 & 69.7 & 71.9 & \textbf{72.5} & 0.59 \\
  + Ind. + on-manifold &  4 & 69.2 & 70.4 & \textbf{72.9} & 70.5 & -- \\
  + Ind. + on-manifold &  8 & 68.8 & 70.0 & 71.5 & 69.1 & -- \\
  + Ind. + on-manifold & 12 & 69.0 & 70.2 & 72.6 & 69.4 & -- \\
  + Ind. + on-manifold & 16 & \textbf{69.3} & 70.3 & 72.4 & 71.5 & -- \\
  \bottomrule
  \end{tabularx}
  \vspace{-4pt}
\end{table}

\begin{figure}[h!]
  \centering
  \includegraphics[width=.5\linewidth]{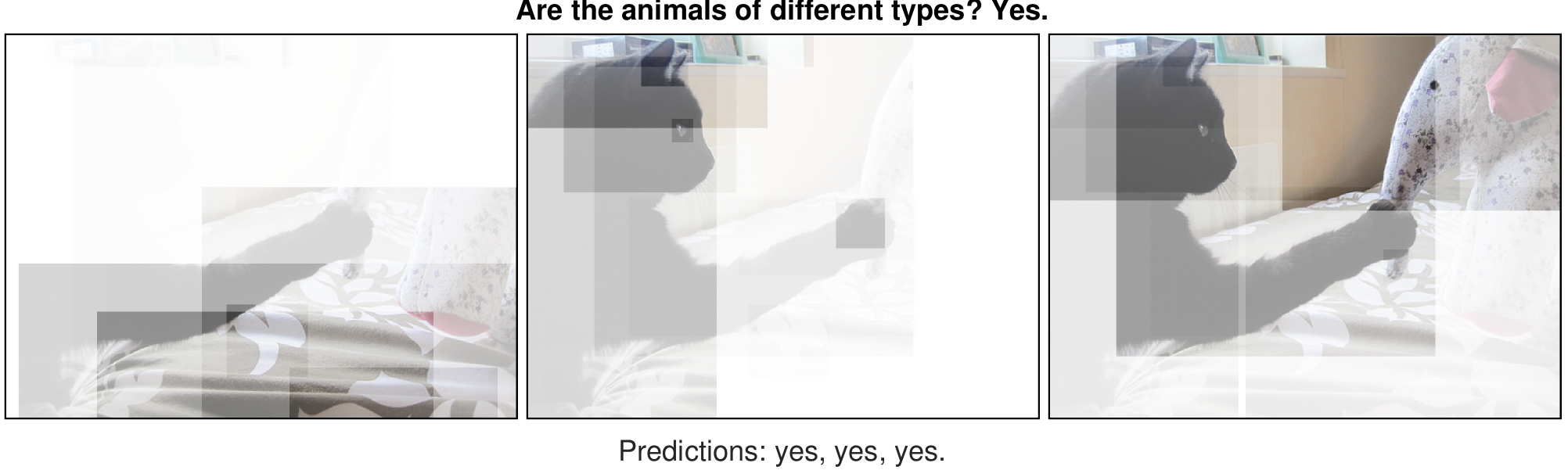}\vspace{3pt}\\
  \includegraphics[width=.5\linewidth]{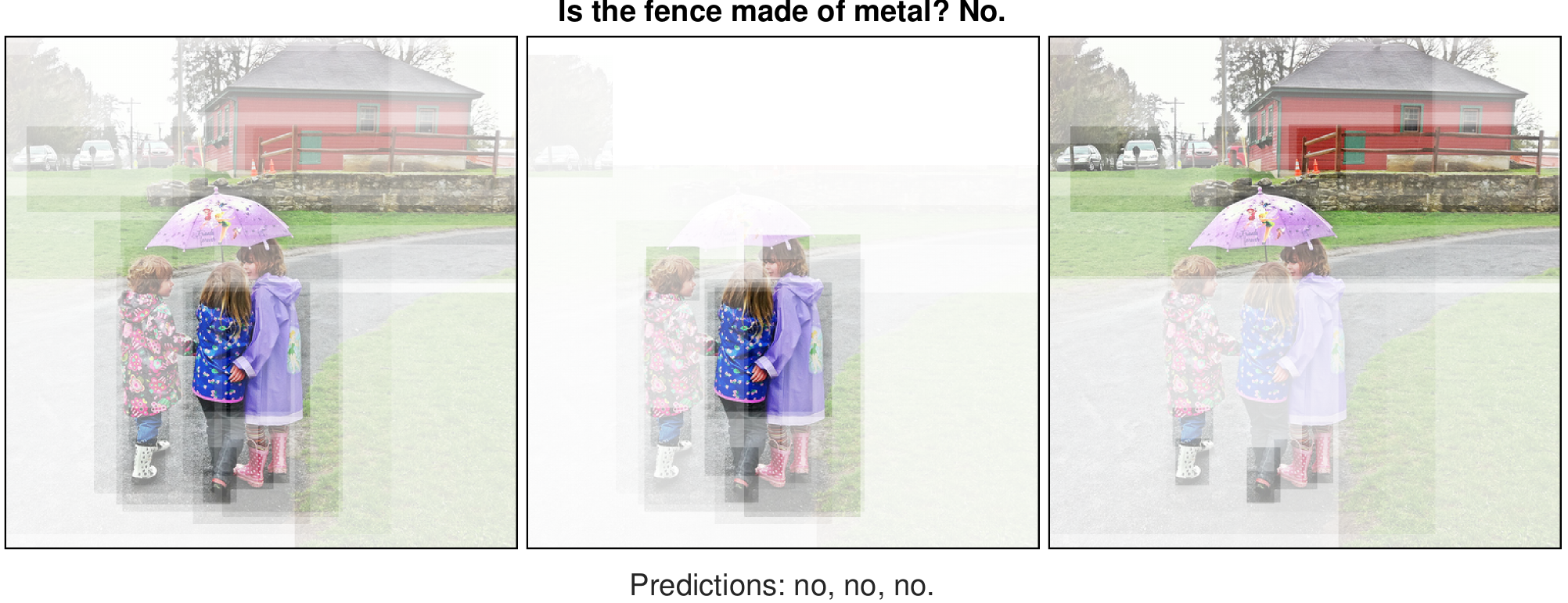}\vspace{3pt}\\
  \includegraphics[width=.5\linewidth]{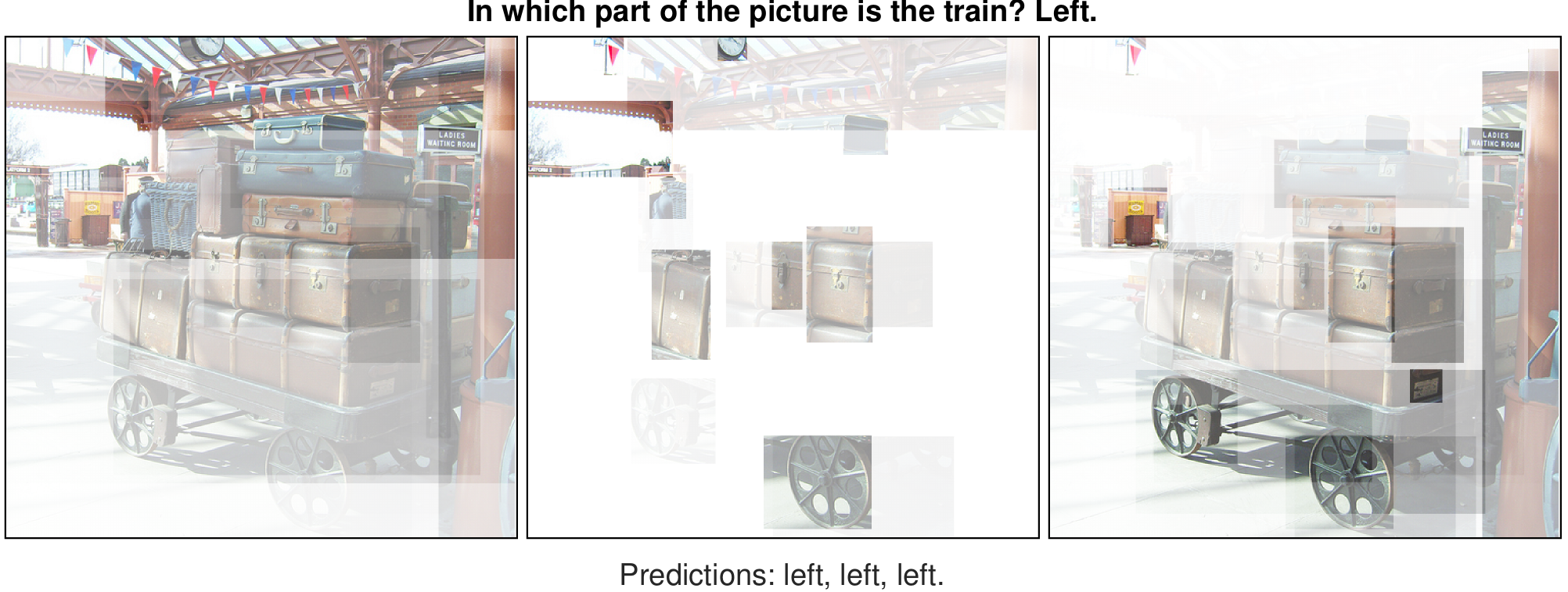}\vspace{3pt}\\
  \vspace{-4pt}
  \caption{
  Examples from the GQA validation set. We show the input question, ground truth answer, and predicted answer from 3 models trained with our method. The input images are weighted with grad-CAM scores over object detections.
  The models often predict the same answer while focusing on different regions.
  This diversity in spatial locations emerges naturally: our independence constraint is applied across the \emph{channels} of globally-pooled visual features.
  \label{figVqaExamples}}
\end{figure}

\clearpage
\section{Discussion}
\label{secDiscussion}

We presented a method that highlights cases of underspecification by training multiple models with similar in-domain performance yet different OOD behaviour.
This method offers a partial solution to building robust models since it discovers features that are otherwise missed by standard ERM due to shortcut learning or other implicit inductive biases~\cite{ortiz2021neural,shah2020pitfalls}.

\myparagraph{What do we gain from identifying cases of underspecification?}
The level of underspecification (indicated by the number of models that can be trained with the proposed constraints) shows how far from unique a solution to a learning problem is.
Diagnosing underspecification is not a pass-or-fail test: all but the simplest tasks and models are underspecified to some extent.
Measuring underspecification should help determining the level of trust attributed to an ML model.
Our constructive approach has the added advantage of exposing the range of predictive features present in the data.


\myparagraph{Importance to both engineering and science.}
There is a continuing source of research questions in the apparent mismatch between empirical practices in ML and some hard limitations of learning methods.
The concept of underspecification has the potential to unify phenomena including shortcut learning, distribution shifts, and adversarial robustness.
These are important for ML as an engineering discipline (improving reliability and applicability of ML methods) as well as a scientific endeavour (understanding the structure of real-world data and how/why existing methods work).

\myparagraph{A first implication} of underspecification is that ERM is insufficient to guarantee OOD generalization.
Identified cases of underspecification point at the need for additional task-specific information in the design of reliable learning methods.
If such information cannot be integrated, learned models are at risk of unexpected behaviour when deployed on OOD data, because the depend on stochastic or arbitrary factors (\eg reliance on texture \vs shape in image classification~\cite{geirhos2018imagenet}).

\myparagraph{A second implication} is that ID and OOD performance are not necessarily coupled.
Without further assumptions, in-domain validation is not a reliable model selection strategy for OOD performance despite contradictory suggestions made in the literature~\cite{gulrajani2020search,miller2021accuracy}.
This strategy might still be useful as a heuristic owing to some inherent structure in real-world data, but its limits of applicability are yet to be understood.

\myparagraph{A third implication} is that high OOD performance of a model is no guarantee for its reliability. High apparent performance might happen by accident in an underspecified setting. In such cases, the model behaviour depends on hidden assumptions and it could still fail unexpectedly.
Identifying underspecification remains important to identify these hidden assumptions, which is
particularly important for high-stakes applications such as medical imaging~\cite{banerjee2021reading,ghimire2020learning}.

The proposed analysis also corroborates existing explanations for techniques that successfully improve generalization, such as data augmentation and contrastive learning.
Both were indeed shown to depend on the injection of additional knowledge, respectively in the design of the augmentations~\cite{cubuk2021tradeoffs,ilse2020designing,von2021self} and pair selection strategy~\cite{zimmermann2021contrastive}.
And this extra knowledge is often task-specific~\cite{xiao2020should}. For example, augmenting images with rotations may help in identifying flowers but not traffic signs.
Injecting task-specific knowledge is sometimes vilified in a ``data-driven'' culture.
This study suggests that we would rather benefit from highlighting this practice and making assumptions more explicit, thus helping one to identify the limits of applicability of various methods.

\myparagraph{Conclusion.} 
This paper made theoretical and methodological steps on the study of underspecification. It complements an observational study~\cite{d2020underspecification} with a method to diagnose and address the problem.

\myparagraph{Limitations.} 
The proposed method for building models with better generalization is only a partial solution since it requires an external model selection procedure.
New methods for model selection~\cite{deng2021does,garg2021ratt,immer2021scalable,wald2021calibration}, robust evaluation~\cite{gardner2020evaluating,kaushik2019learning}, and explainability~\cite{goyal2019counterfactual,thiagarajan2021designing} are all suitable to implement this selection.
Interactive approaches~\cite{das2019beames} are another option that injects expert knowledge.
Another possible extension is to apply the method to the end-to-end training of larger models.
Finally, this work focused on i.i.d. training data.
We hope to extend the analysis to forms of data known to be valuable for OOD generalization such as multiple environments~\cite{arjovsky2019invariant,peters2016causal,teney2020unshuffling}, counterfactual examples~\cite{kaushik2019learning,teney2020learning}, and non-stationary data~\cite{alesiani2021gated,halva2020hidden,pfister2019invariant,venkateswaran2021environment}.
The analysis of multi-environment training, as used in domain generalization, may elucidate why these methods are often ineffective in practice~\cite{gulrajani2020search}.

\bibliographystyle{splncs04}
\bibliography{0-main}

\clearpage
\appendix
\section*{Appendices}

\floatsetup[figure]{capposition=below,floatwidth=\textwidth}
\floatsetup[table]{capposition=below,floatwidth=\textwidth}

\section{Analogy with disentanglement in representation learning}
\label{disentanglement}

There are parallels between our discovery of simple predictive features and the topic of disentanglement in representation learning.
Disentanglement aims to identify all independent factors of variation in the data-generating process~\cite{bengio2013representation} \ie in a \textbf{task-independent} manner.
In comparison, we are interested only in features that are correlated with \textbf{task-specific} labels in a given dataset.
Many disentanglement methods are based on the independence of the factors (even though it is usually insufficient~\cite{trauble2021disentangled}) while we rely on the independence of \emph{predictions} with each feature.
Our method can therefore be seen as a form of disentanglement in the space of predictors.
This parallel also has implications for identifiability.
Disentanglement of the correct generative factors was shown to be impossible without additional assumptions or supervision~\cite{locatello2019challenging}.
In our setting, the identification of features that are causally-related to the label (as opposed to spuriously-correlated) also requires additional information~\cite{bareinboim2020pearl,scholkopf2021toward}.
Our formulation relegates this need to an external ``model selection strategy'' that allows flexibility in its actual realization.

\section{Projections on the manifold}
\label{appendixProjection}
The on-manifold loss defined in Eq.~(\ref{lossOnManifold}) requires projecting a model's input gradients on the data manifold.
The manifold is characterized by a provided set of unlabeled data $\dataUnlabelled = \{\bx_i\} \sim \probOod$.
We propose two implementations of the projection function $\project(\cdot)$, using either a simple principal component analysis (PCA) model of the manifold or a variational auto-encoder (VAE).
A PCA is fast to evaluate but can only model a linear manifold, which is likely overly simplistic for most real datasets. A VAE allows learning a non-linear manifold.    
Better implementations are probably possible by taking advantage of recent developments in OOD detection and generative models such as GANs, EBMs, and diffusion models.

In the case of a PCA model, the projection is a straightforward linear projection on the top components of a PCA basis of $\dataUnlabelled$.
In the case of a VAE, we train an auto-encoder on the data $\dataUnlabelled$. We interpret it as a generative model of the manifold, assuming that inputs to the auto-encoder will be mapped to a nearby projection on the manifold.
We denote the auto-encoder as a function
$\aenc_\bpsi:~\mathbb{R}^\din  \rightarrow \manifold$
of parameters $\btheta$ such that
\abovedisplayskip=4pt \belowdisplayskip=4pt
\begin{align}
  \aenc_\bpsi(\bx) \smallApprox \bx \quad\forall~\bx\smallSim\probOod.
\end{align}
Therefore, the reconstruction error $||\bx - \aenc_\bpsi(\bx)||$ is minimal for points $\bx$ on to the manifold.

The function $\aenc$ is trained to reconstruct points, but it is also readily capable of projecting a vector $\deltax \in \mathbb{R}^\din$ originating at $\bx$.
In our case, we will use it to project the input gradient ($\deltax\smallEq\nabla_\bx f\btheta(\bx)$).
We overload the function as $\aenc_\bpsi: \mathbb{R}^\din \times \mathbb{R}^\din \,\rightarrow\, \manifold$ with
\begin{align}
  \aenc_\bpsi(\bx, \deltax) \smallApprox \deltax \quad\forall~\bx  \smallSim \probOod, ~~(\bx+\deltax) \smallSim \probOod.
\end{align}
Here again, the reconstruction error $||\deltax - \aenc_\bpsi(\bx, \deltax)||$ is minimal for a point $\bx$ on the manifold and vector $\deltax$ \textbf{aligned with the manifold}.
We can therefore apply $\project(\cdot)$ to our input gradients, measure the distance with their projection on the manifold, and use this distance as our on-manifold loss in Eq.~(\ref{lossOnManifold}).

\myparagraph{Reconstructing vectors with an auto-encoder.}
\label{autoEncGradients}
This section describes how to take a standard auto-encoder, typically trained to reconstruct a point~$\bx$ of its input space, and use it to reconstruct a vector~$\deltax$ in this space (input gradients in our case).
The auto-encoders used this work are compositions of linear layers, ReLU activations, and a sigmoid output activation.
Table~\ref{tabVaeGradients} provides the equivalent operations performed during forward propagation in any such layer for reconstructing an input or gradient.

\noindent
\begin{table}[h!]
  \centering
  \caption{Operations performed at each layer of a VAE during forward propagation for reconstructing points (left column) and vectors/gradients~(right column).\label{tabVaeGradients}}
  \vspace{6pt}
  \small
  \renewcommand{\tabcolsep}{10pt}
  \renewcommand{\arraystretch}{1.0}
  \begin{tabular}{lll}
    \toprule
    \textbf{Operation on point $\bx$} & ~ & \textbf{Operation on vector $\deltax$}\\
    \midrule
    \multicolumn{3}{l}{Linear layer}\\
    $\bx \,\leftarrow\, \bW \bx \,+\, \bb$ & ~ & $\deltax \,\leftarrow\, \bW \deltax$\\
    \midrule
    \multicolumn{3}{l}{ReLU activation}\\
    $\bx \,\leftarrow\, \bx \,\odot\, \indicatorSmall(\bx > 0)$ & ~ & $\deltax \,\leftarrow\, \deltax \,\odot\, \indicatorSmall(\bx > 0)$\\
    \midrule
    \multicolumn{3}{l}{Sigmoid activation}\\
    $x \,\leftarrow\, \sig(x) = 1 / (1+e^{-x})$ & ~ & $\deltax \,\leftarrow\, \deltax \; \sig(x) \; \big(1-\sig(\bx)\big)$\\
    \bottomrule
  \end{tabular}
\end{table}

\section{Distilling multiple models into one}

After training a set of models, we propose to combine the best of them into one a global predictor with superior OOD performance.
We train this predictor with the same fine-tuning as described in~\secref{secFineTuning}, using masks of selected models aggregated with a logical OR.
We outline in \algref{algCombining} a simple procedure to iteratively combine models pairwise and greedily. 

\begin{algorithm}[h!]
  \small
  \caption{Distilling multiple models into one with greedy pairwise combinations.}
  \label{algCombining}
  \DontPrintSemicolon
  \BlankLine
  \KwInput{\\
    $\bS = \{ f_{\btheta_1} \, \mydots \, f_{\btheta_M}\}$: Set of independent models.\\
    $\operatorname{getNBestModels}(\cdot)$: Model selection strategy, \eg evaluation on OOD validation set.
  }
  \BlankLine
  \KwOutput{\\
    $f_{\btheta_\star}$: Best combined model according to given strategy.
  }
  \BlankLine
  \KwResult{}
  $\wtIndep \leftarrow 0, ~\wtManifold \leftarrow 0$~~\myComment{Use only predictive loss}\\
  \Do{$\operatorname{get1BestModel}(\bS) = \btheta_\star$ ~\myComment{New combination is best}}{
    $\{ \btheta_{k}, \btheta_{l} \} \leftarrow \operatorname{get2BestModels}(\bS)$\\ 
    $\mask^\star_i \leftarrow \mask^{k}_i \,\lor\, \mask^{l}_i ~~~\forall i$~~\myComment{Combine masks}\\
    $\dataTr^\star \leftarrow \{(\bx_i \odot \mask^\star_i, y_i)\}_i$~~\myComment{Prepare masked data}\\
    $\btheta_\star \leftarrow \argmin \mathcal{L}(\dataTr^m, \btheta_{k})$~~\myComment{Fine-tune}\\
    $\bS \leftarrow \bS ~\bigcup~ \btheta_\star$~~\myComment{Append to set of models}\\
  }
  $\btheta_\star ~\leftarrow~ \operatorname{get1BestModel}(\bS)$~~\myComment{Return best model}
  \BlankLine
\end{algorithm}

\section{Experimental details: collages}
\label{appendixImplementation}

\myparagraphWoSpacing{Collages dataset, raw pixels as input.}
The \emph{collages} dataset is described in~\cite{teney2021evading}.
We used the {4-block ordered version} (each of the four source datasets appear in the same quadrant in every instance).
We re-generated the collages with {bilinear downsampling by 1/4th} using the code from the authors.%
\footnote{\url{https://github.com/dteney/collages-dataset}}
The reason we do not use full-size images is purely computational.
The original data~\cite{teney2021evading} used nearest-neighbour downsampling because it preserves the contrast and dynamic range,
but bilinear downsampling produces smoother images that better match the manifold assumption of natural data that is important in this work.
The bilinear downsampling causes a compression of the dynamic range that is uneven across the tiles and mostly affects SVHN and CIFAR.
To compensate for it, we apply a standard local contrast normalization ({\small\texttt{adapthisteq}} in Matlab).
We also replace the pixels with a constant value near 0 with small noise. These pixels exist in the MNIST data but are again unrepresentative of natural data.

\begin{minipage}{\textwidth} 
\vspace{5pt}
\noindent
In Matlab code:
\begin{mdframed}[style=codeFrame]
  \scriptsize
  \texttt{
    \textcolor{darkGreen}{\% Local contrast normalization}\\
    img = adapthisteq(img, 'NumTiles', [2 2]);\vspace{3pt}\\
    \textcolor{darkGreen}{\% Replace dark pixels with small random values}\\
    img(img < .05) = rand(nnz(img < .05), 1) * .05;
  }
\end{mdframed}
\end{minipage}

\myparagraph{Collages dataset, ResNet features as input.}
We obtain feature maps by passing full-resolution images of collages into a standard frozen ImageNet-pretrained ResNet-18~\cite{he2015resnet}.
We chose to use feature maps from an intermediate layer {\small\texttt{res3b\_relu}} because it retains a reasonable spatial resolution (8$\times$8 with 128 channels).
The independence loss (\ref{lossIndep}) for these experiments uses a dot product rather than a cosine similarity.
It works better but we don't know why.

\myparagraph{Hyperparameters.}
See \tabref{tabHyperparametersCollages}.
The classifiers' hyperparameters are chosen to optimize the baseline reported as ``Upper bounds'' in the results.
None of these hyperparameters are particularly tuned to the proposed methods.

\begin{table*}[p!]
  \centering
  \renewcommand{\tabcolsep}{3pt}
  \renewcommand{\arraystretch}{1.02}
  \caption{Hyperparameters used on {collages} (top), GQA, and WILDS-Camelyon17 (bottom).\label{tabHyperparametersCollages}\label{tabHyperparametersVqa}}\vspace{8pt}
  \scriptsize
  \begin{tabularx}{\linewidth}{Xllll}
    \toprule
    \textbf{Hyperparameter} & ~ & \textbf{Collages, pixels as input} & ~ & \textbf{ResNet features as input}\\
    \midrule
    \textbf{Classifier}&&\\
    Input dimensions && 16$\times$16$\times$1 ~(grayscale) && 8$\times$8$\times$128 ~~(res3b\_relu layer)\\
    Architecture && 1 Hidden layer (fully-connected) && 1 Hidden layer (channel-wise)\\
    ~ && 8 neurons && 16 neurons\\
    && Output layer (fully-connected). && Output layer (fully-connected)\\
    Hidden Activations && Leaky ReLU, leak scale 0.01 && Leaky ReLU, leak scale 0.01\\
    Output activation && Sigmoid && Sigmoid\\
    Mini-batch size && 256 && 256\\
    Optimizer && Adam && Adam\\
    Learning rate && 0.002 && 0.001\\
    Optimization length && 10,000 Updates && 30,000 Updates\\
    ~ && No early stopping && No early stopping\\
    \midrule
    \textbf{PCA Manifold model}&&\\
    Number of components && 24 && --\\
    Retained variance && 85\% && --\\
    \midrule
    \textbf{VAE Manifold model}&&\\
    Architecture && 2-Hidden layer MLP && 2-Hidden layer MLP\\
    ~ && 128 neurons per layer && 128 neurons per layer\\
    Latent dimensions && 24 && 24\\
    Hidden activations && ReLU && ReLU\\
    Output activations && Sigmoid && Sigmoid\\
    Mini-batch size && 256 && 128\\
    Optimizer && Adam && Adam\\
    Learning rate && 0.001 && 0.001\\
    Optimization length && 100 Epochs && 300 Epochs\\
    Weight of KL loss && 0.01 && 0.01\\
    ~ && \multicolumn{3}{l}{(small to prioritize accurate reconstructions)}\\
    \bottomrule

    ~\\~\\~\\~\\
    \toprule
    \textbf{Hyperparameter} & ~ & \textbf{GQA} & ~ & \textbf{WILDS-Camelyon17}\\
    \midrule
    \textbf{Classifier}&&\\
    Input dimensions && 1$\times$300~(text) && 1$\times$1024~(DenseNet features)\\
    ~ && 1$\times$2048~(image) && \\
    Hidden layers dimension && 256 && --\\
    Hidden activations && Leaky ReLU && --\\
    Output activations && Sigmoid && Sigmoid\\
    Mini-batch size && 256 && 256\\
    Optimizer && Adam && Adam\\
    Learning rate && 0.002 && 0.001\\
    Optimization length && 50,000 Updates (40 epochs) && 12,000 Updates (10 epochs)\\
    ~ && No early stopping && \\
    \midrule
    \textbf{PCA Manifold model}\\
    Number of components && 168 (chosen for best performance) && Same as number of models\\
    Retained variance && 85\% && Varies: 75\% with 2 components, 85\% with 3, \\
    ~ && && 90\% with 6, 95\% with 13, 97\% with 25\\
    \midrule
    \textbf{VAE Manifold model}&& && \\
    Architecture && 2-Hidden layer MLP && 1-Hidden layer MLP\\
    ~ && 512 neurons per layer && 128 neurons per layer\\
    Latent dimensions && 32 && 14\\
    Hidden Activations && ReLU && ReLU\\
    Output activation && None && None\\
    Mini-batch size && 256 && 256\\
    Optimizer && Adam && Adam\\
    Learning rate && 0.001 && 0.001\\
    Optimization length && 100 Epochs && 20 Epochs\\
    Weight of KL loss && 0.01 && 0.01\\
    \bottomrule
  \end{tabularx}
\end{table*}

\myparagraph{Baselines.}
The ``dropout'' baseline in \tabref{tabCollagesPixels} is implemented with as a standard dropout on the input pixels, with a different dropout mask for each model. The idea is that each model sees a different (dropped-out) version of the input, and therefore might pick up different features.

\section{Experimental details: GQA}
\label{appendixVqa}

\myparagraphWoSpacing{VQA Model.}
For our experiments on visual question answering, we use a simplified version of the classical BUTD model~\cite{teney2017challenge}. For the text input (question), we use standard 300-dimensional GloVE embeddings~\cite{pennington2014glove} (frozen) averaged over the sequence. For the image input, we use the object features provided with the GQA dataset. These are 2048-dimensional from a ``bottom-up'' object detector~\cite{anderson2017features}. We average and L2-normalize these features to obtain a single 2048D vector representing each image.
The text and question vectors are linearly projected to a common dimension (256), combined with an element-wise product, then passed through a 1-layer MLP to obtain scores over candidate answers.

\myparagraph{Grad-CAM Rank correlation.}
The rank correlation reported in \tabref{tabVqa} is the Spearman rank correlation of grad-CAM scores on the validation set, average over questions and pairs of models. The grad-CAM scores correspond to the 2048D input gradients multiplied by the unpooled (36 $\times$ 2048D) object features. The grad-CAM scores are used because they provide a spatial importance map over the image despite the model using globally-pooled features (hence an input gradient that is spatially uniform over the image).

\noindent
The proposed method is applied in the same way as with the collages.
Almost all hyperparameters are remarkably identical to those for the collages (see \tabref{tabHyperparametersVqa}).
\clearpage
\onecolumn
\section{Additional results: collages}

\noindent
We provide below additional results on the \emph{collages} dataset.
\label{appendixCollagesResNet}
We also include experiments using \textbf{features from a ResNet-18 as inputs} rather than raw pixels.
We use a standard ResNet-18 pretrained on ImageNet then kept frozen.
Our method is applied identically as in other experiments, except that the independence and on-manifold constraints are now applied in the space of ResNet feature maps rather than in pixel space.
The manifold model is a VAE trained on such feature maps.
\textbf{The behaviour of our method with ResNet features is qualitatively the same as in other experiments.}
This demonstrates the applicability of the method on features from deep architectures.

\begin{figure*}[h!]
  \centering
  \begin{subfigure}[t]{0.18\linewidth}
    \centering\includegraphics[width=0.90\linewidth]{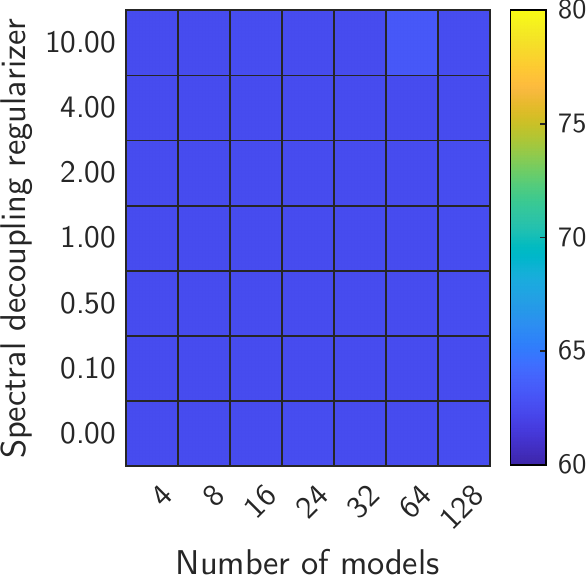}
    \caption{\tiny Spectral decoupling.}
  \end{subfigure}
  \begin{subfigure}[t]{0.18\linewidth}
    \centering\includegraphics[width=0.90\linewidth]{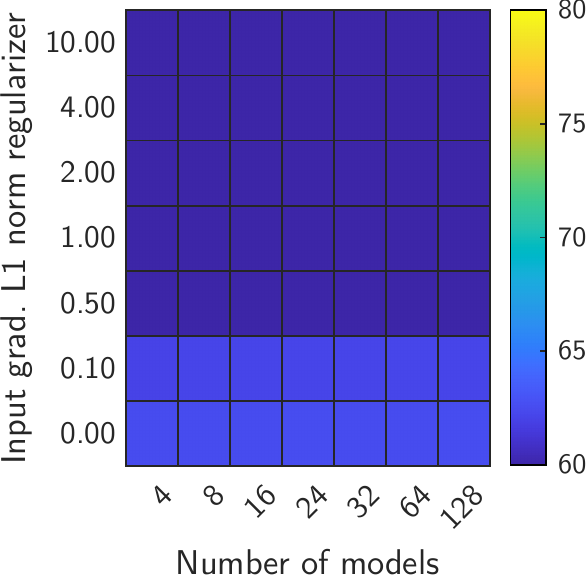}
    \caption{\tiny Input grad. L1 norm.}
  \end{subfigure}
  \begin{subfigure}[t]{0.18\linewidth}
    \centering\includegraphics[width=0.90\linewidth]{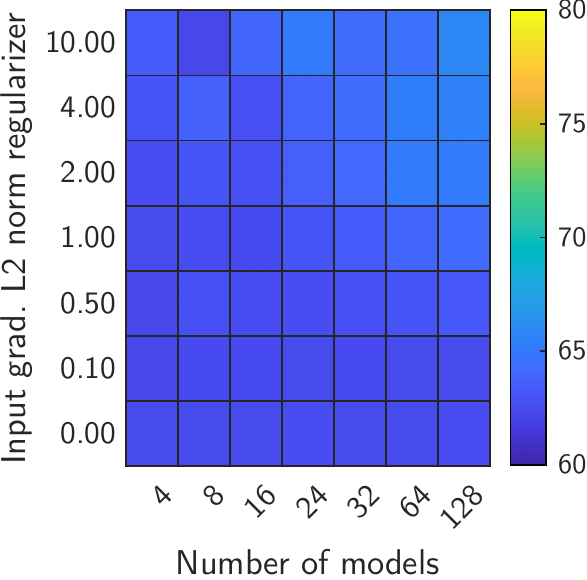}
    \caption{\tiny Input grad. L2 norm.}
  \end{subfigure}
  \begin{subfigure}[t]{0.18\linewidth}
    \centering\includegraphics[width=0.90\linewidth]{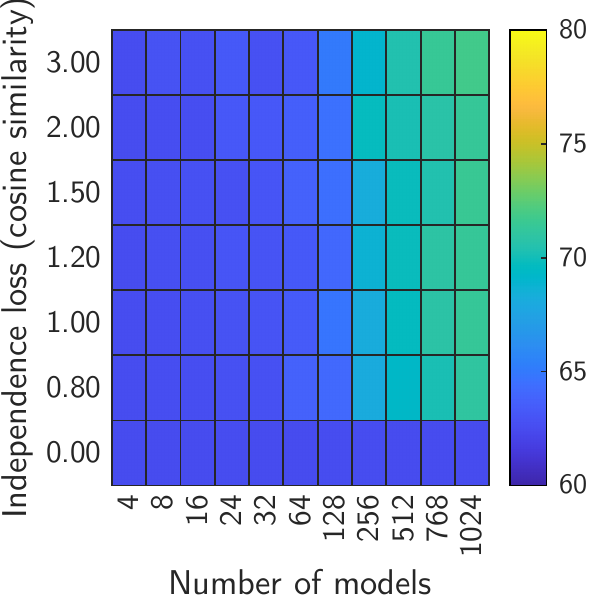}
    \caption{\tiny Indep. loss (cos. sim.)}
  \end{subfigure}
  \begin{subfigure}[t]{0.18\linewidth}
    \centering\includegraphics[width=0.90\linewidth]{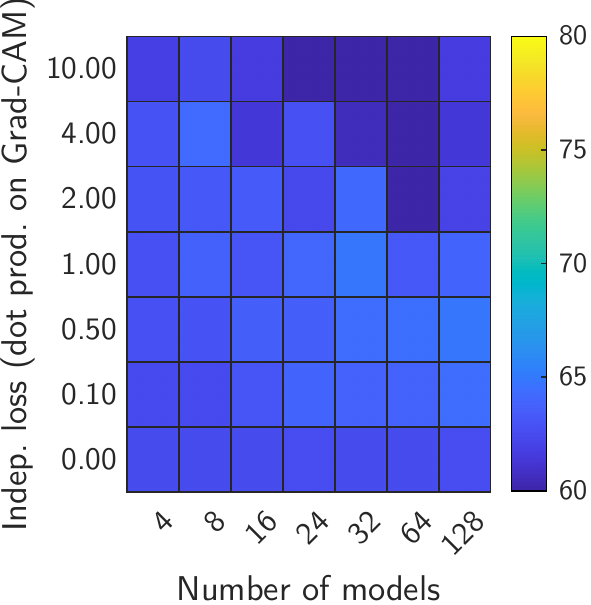}
    \caption{\tiny Dot prod. on Grad-CAM.}
  \end{subfigure}
  \vspace{13pt}\\
  \begin{subfigure}[t]{0.18\linewidth}
    \centering\includegraphics[width=0.90\linewidth]{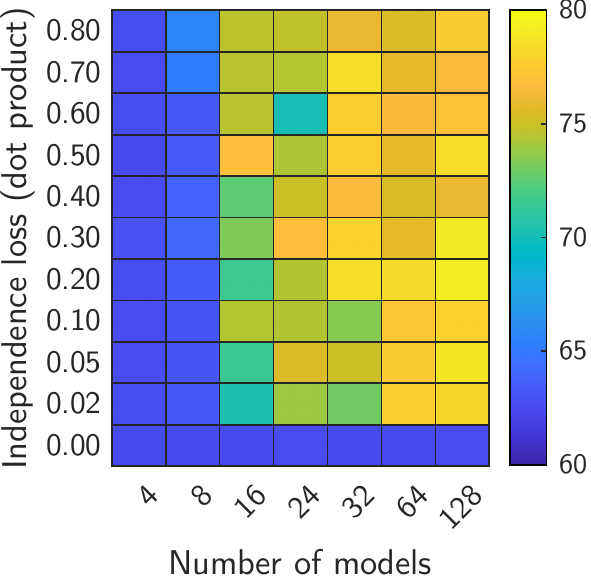}
    \caption{\tiny Indep. loss (dot prod.)}
  \end{subfigure}
  \begin{subfigure}[t]{0.18\linewidth}
    \centering\includegraphics[width=0.90\linewidth]{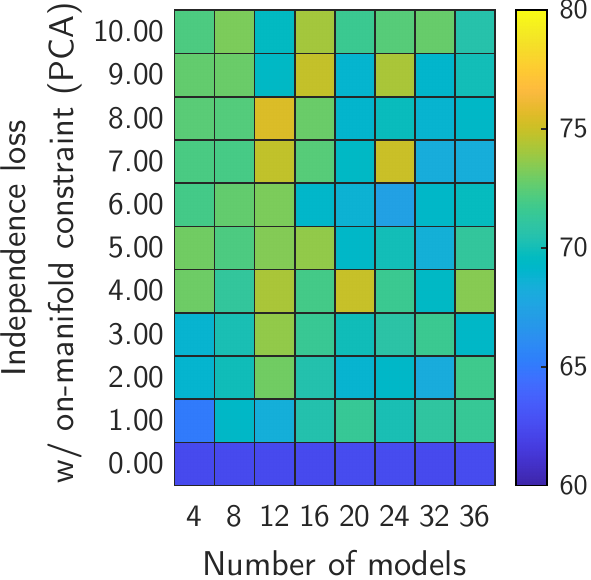}
    \caption{\tiny Indep. loss (cos. sim.) + on-manifold (PCA).}
  \end{subfigure}
  \begin{subfigure}[t]{0.18\linewidth}
    \centering\includegraphics[width=0.90\linewidth]{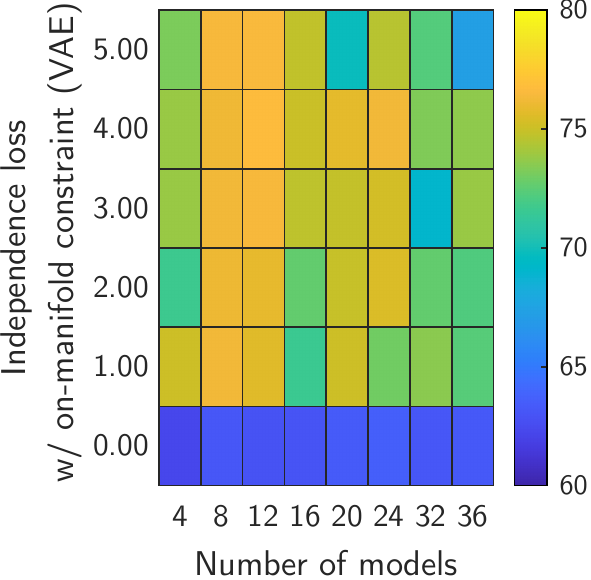}
    \caption{\tiny Indep. loss (cos. sim.) + on-manifold (VAE).}
  \end{subfigure}
  \begin{subfigure}[t]{0.18\linewidth}
    \centering\includegraphics[width=0.90\linewidth]{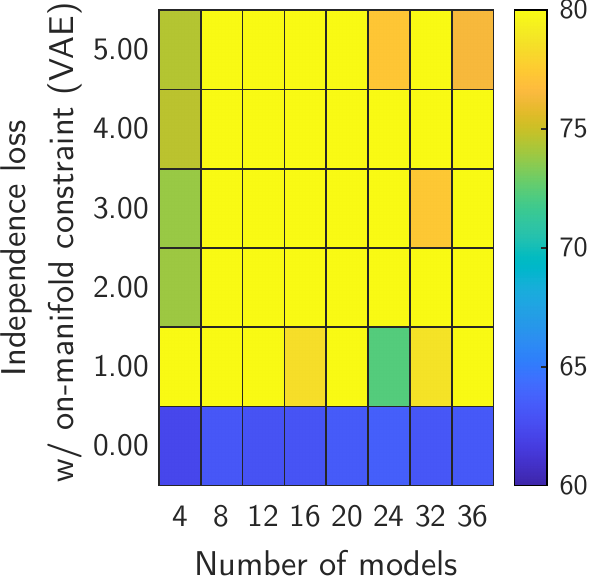}
    \caption{\tiny Indep. loss (cos. sim.) + on-manifold (VAE) + fine-tuning.}
  \end{subfigure}
  \begin{subfigure}[t]{0.18\linewidth}
    \centering\includegraphics[width=0.90\linewidth]{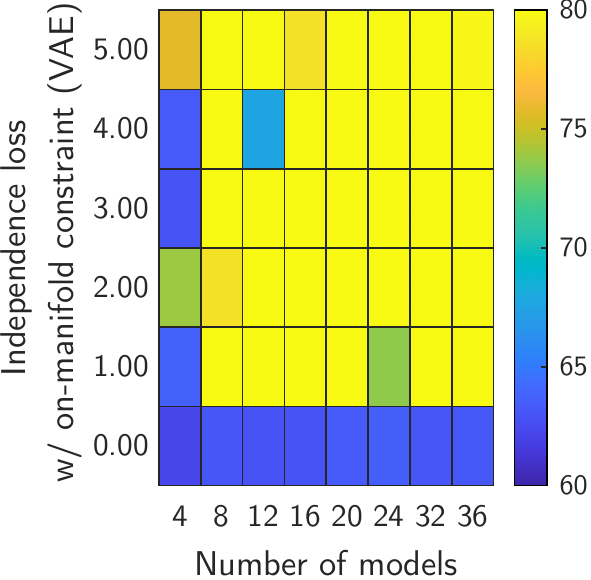}
    \caption{\tiny Indep. loss (cos. sim.) + on-manifold (VAE) + fine-tuning + combinations.}
  \end{subfigure}
  \caption{\textbf{(Collages, pixels as input)}~
  Accuracy (average over the 4 test sets) of \textbf{existing methods (first row)} and \textbf{ablations (second row)} for various hyperparameters and numbers of models.
  The only existing methods with non-trivial performance (\eg (f)) require training at least an order of magnitude more models than ours (j).
  \label{figCollagesHeatmapsAllMethods}}
\end{figure*}

\begin{figure}[h!]
  \centering
  \begin{subfigure}[t]{0.4\linewidth}
    \centering\includegraphics[width=0.6\linewidth]{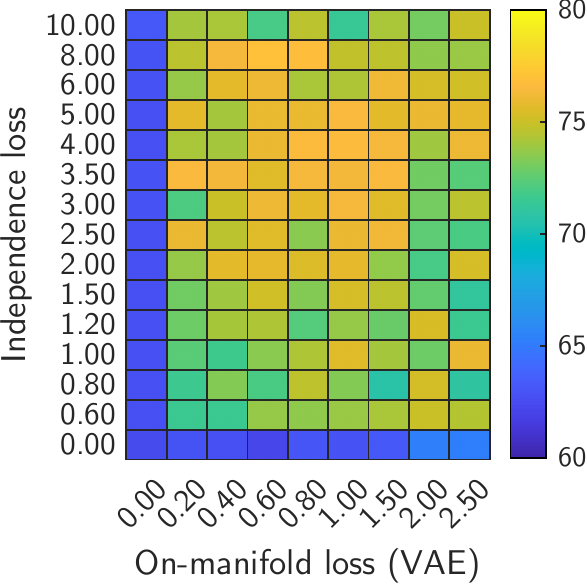}\vspace{6pt}
    \caption{Pixels as input.}
  \end{subfigure}
  \begin{subfigure}[t]{0.4\linewidth}
    \centering\includegraphics[width=0.6\linewidth]{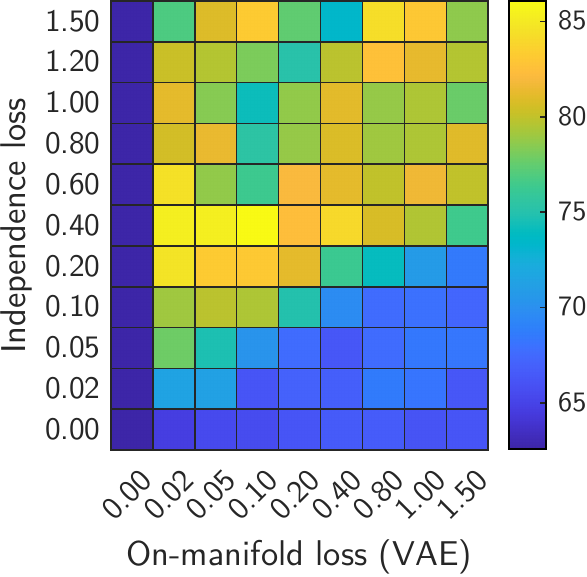}\vspace{6pt}
    \caption{ResNet features as input.}
  \end{subfigure}
  \vspace{0pt}\\
  \caption{\textbf{(Collages)}~
  Accuracy of the proposed method (average over the 4 test sets) for various loss weights of the independence and on-manifold losses (with 12 models).
  \textbf{Performance is stable over a range of values} and both losses are important, as seen from the leftmost \& lowermost cells.
  Performance is also repeatable: each cell reports a single run \ie {not} averaged over multiple random seeds.
  \label{figCollagesHeatmaps2}}
\end{figure}

\begin{figure}[h!]
  \centering
  \begin{subfigure}[t]{0.3\linewidth}
    \centering\includegraphics[width=0.7\linewidth]{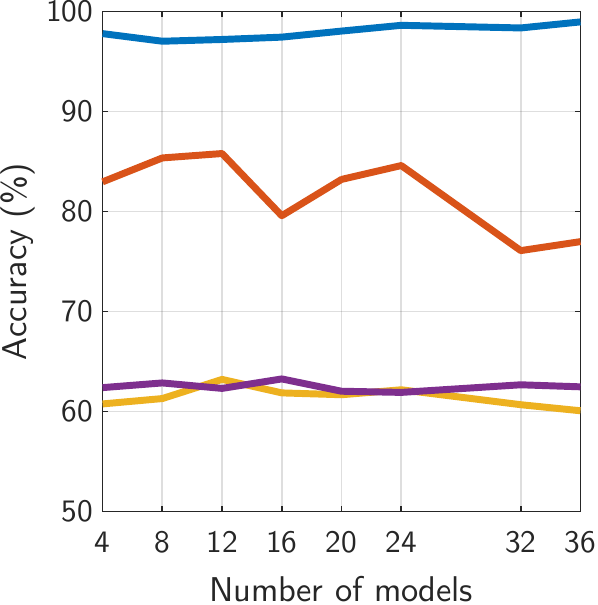}\vspace{6pt}
    \caption{Pixels as input.}
  \end{subfigure}
  \begin{subfigure}[t]{0.3\linewidth}
    \centering\includegraphics[width=0.7\linewidth]{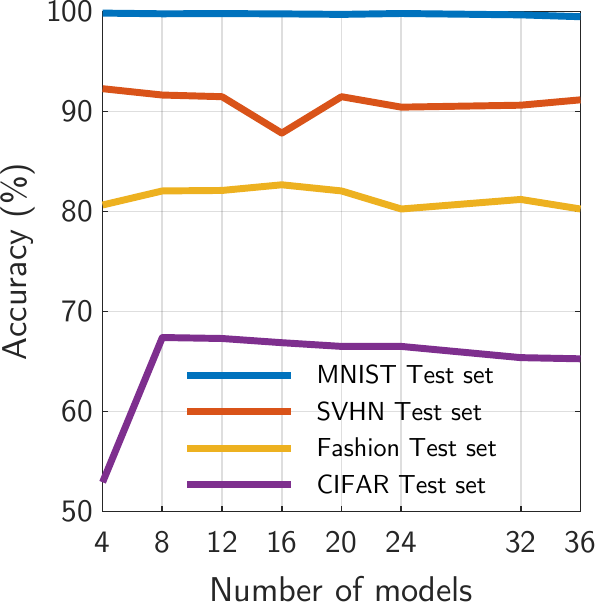}\vspace{6pt}
    \caption{Pixels as input + fine-tuning.}
  \end{subfigure}
  \begin{subfigure}[t]{0.3\linewidth}
    \centering\includegraphics[width=0.7\linewidth]{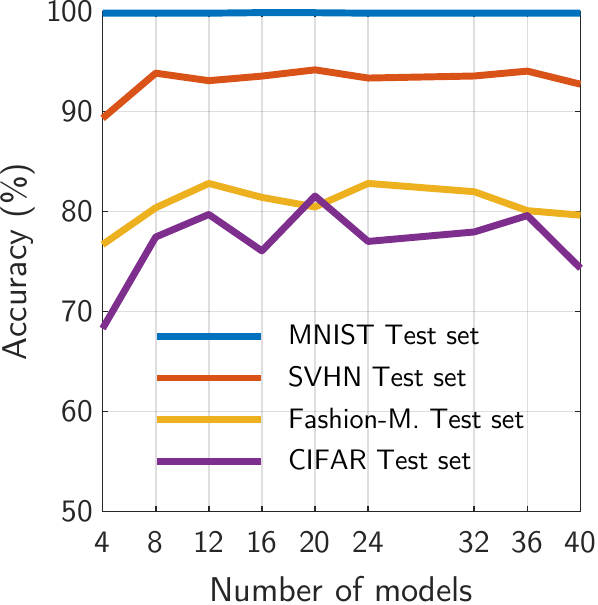}\vspace{6pt}
    \caption{ResNet features as input.}
  \end{subfigure}
  \caption{\textbf{(Collages)}~
  Accuracy of the proposed method as a function of the number of models trained.
  Non-trivial accuracy is obtained on all 4 test sets with as few as 4 models.
  In line with our theoretical predictions, the accuracy decreases after $>$24 models, which is the recommended value based on the intrinsic dimensionality of the dataset (\secref{secIntrinsicDimensionalityEstimation}).
  \label{figCollagesNumModels}}
\end{figure}

\begin{figure*}[h!]
  \centering
  \begin{subfigure}[t]{1\linewidth}
    \centering\includegraphics[width=1\linewidth]{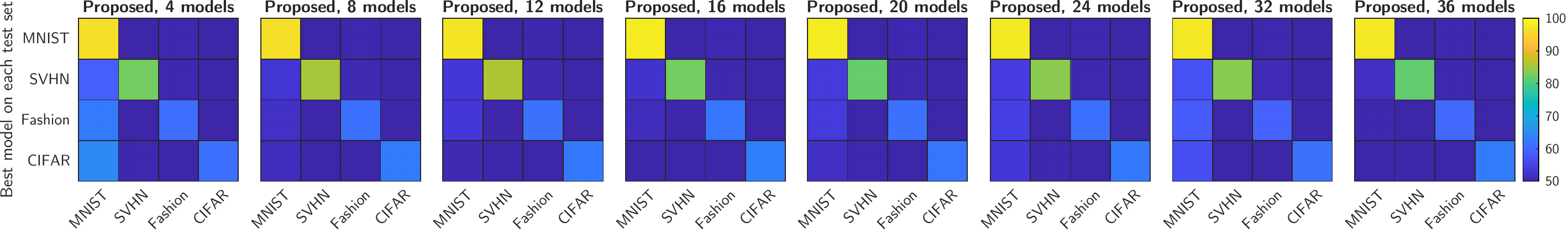}\vspace{3pt}
    \caption{\textbf{Pixels} as input.}
  \end{subfigure}
  \vspace{3pt}
  \\
  \begin{subfigure}[t]{1\linewidth}
    \centering\includegraphics[width=1\linewidth]{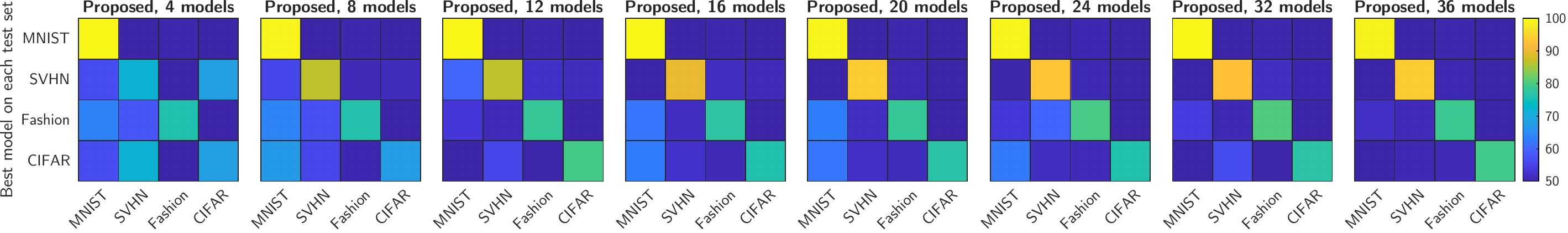}\vspace{3pt}
    \caption{\textbf{ResNet} features as input.}
  \end{subfigure}
  \\
  \caption{\textbf{(Collages)}~
  Accuracy on the 4 test sets~(columns) of models with the best accuracy on each set~(rows).
  The diagonal pattern indicates that the models specialize and learn different, non-overlapping features.
  \label{figCollagesMatrices2}}
\end{figure*}

\begin{table}[t]
  \scriptsize
  \renewcommand{\tabcolsep}{0.35em}
  \renewcommand{\arraystretch}{1.02}
  \centering
  \begin{tabularx}{1.0\linewidth}{Xccccc}
  \toprule
  \textbf{Collages} dataset, \textbf{ResNet} features as input (accuracy in \%) & \multicolumn{4}{c}{Best model on} \vspace{1.5pt}\\ \cline{2-5}
  ~ &  \rotatebox{90}{MNIST} & \rotatebox{90}{SVHN~~} & \rotatebox{90}{Fashion\phantom{e}} & \rotatebox{90}{CIFAR-10\phantom{e}} & \rotatebox{90}{Average}\\
  \midrule
  Upper bounds (training on OOD data) & 100.0 & 98.0 & 92.3 & 89.9 & 95.1 \\
  \midrule
  Baseline & 99.8 & 49.9 & 50.7 & 49.9 & 62.6 \\ 
  \midrule
  Independence loss\\
  4 models & 86.8 & 60.2 & 62.1 & 51.6 & 65.2 \\ 
  8 models & 88.4 & 61.5 & 63.3 & 51.9 & 66.3 \\ 
  16 models & 91.6 & 59.3 & 59.9 & 52.1 & 65.7 \\ 
  24 models & 93.9 & 58.2 & 57.7 & 52.1 & 65.5 \\ 
  \midrule
  Independence loss + on-manifold constraint (VAE)\\
  4 models & 99.6 & 71.7 & 76.1 & 68.3 & 78.9 \\ 
  8 models & 99.2 & 86.8 & 76.3 & 67.8 & 82.5 \\ 
  16 models & 99.5 & 90.0 & 77.5 & 76.1 & 85.8 \\ 
  \textbf{24 models} & \textbf{99.2} & \textbf{93.0} & \textbf{80.1} & \textbf{76.0} & \textbf{87.1} \\ 
  \bottomrule
  \end{tabularx}
  \vspace{4pt}
  \normalsize
  \caption{
    Accuracy of the proposed method and ablations using ResNet features as input.
    The independence and on-manifold constraints are both important for high performance.
    There remains a small gap with the upper-bound accuracy, but we achieve drastically better results than the baseline.
    \label{tabCollagesResnet}}
\end{table}

\clearpage
\onecolumn
\section{Additional results: WILDS-Camelyon17}
\label{appendixWilds}

\begin{table}[h!]
  \caption{Best accuracy (\%) on the test-OOD split of WILDS-Camelyon17 (average over 6 random seeds). We show the difference in performance between the baseline and various ablations of our method while training 12 models (as in \tabref{tabCamelyon}), starting with each of the 10 pretrained models provided by the dataset authors~\cite{koh2020wilds}. \textbf{While the ranking of methods is roughly similar, the absolute accuracy (first and last rows) is highly variable across pretrained models.}
  A possible solution for eliminating this variability would be to apply our method during pretraining --~at far greater computational expense than when training linear classifiers.\label{tabCamelyonAppendix}}
  \small
  \renewcommand{\tabcolsep}{0.4em}
  \renewcommand{\arraystretch}{1.1}
  \centering
  \begin{tabularx}{\linewidth}{Xrrrrrrrrrr}
    \toprule
    \textbf{WILDS-Camelyon17} & \multicolumn{10}{c}{Pretrained model}\\
    ~ & \#1~ & \#2~ & \#3~ & \#4~ & \#5~ & \#6~ & \#7~ & \#8~ & \#9~ & \#10~ \\
    \midrule
    ERM Baseline                                 & 68.4 & 78.0 & 73.3 & 60.4 & 78.3 & 78.5 & 64.4 & 74.2 & 71.9 & 60.2 \\
    ~ & \pp{0.1}~ & \pp{0.1}~ & \pp{0.0}~ & \pp{0.1}~ & \pp{0.0}~ & \pp{0.1}~ & \pp{0.1}~ & \pp{0.0}~ & \pp{0.2}~ & \pp{0.1}~ \\
    \midrule
    + Independence constraint                    & +6.2 & +4.5 & +0.8 & +4.4 & -1.5 & +2.9 & +1.3 & +0.8 & +4.1 & +2.3 \\
    + On-manifold soft regularizer, VAE          & +11.9 & \textbf{+7.1} & +0.6 & +11.5 & -1.0 & +5.1 & +3.6 & +2.2 & +6.2 & +2.3 \\
    + On-manifold hard projection, VAE           & +7.9 & +4.2 & +2.2 & +12.9 & -0.3 & \textbf{+7.4} & \textbf{+9.9} & +2.6 & +7.1 & \textbf{+4.5} \\
    + On-manifold soft regularizer, PCA          & +10.6 & +4.7 & +2.6 & +8.6 & -0.4 & +6.4 & +7.4 & +2.9 & \textbf{+8.4} & +4.2 \\
    + On-manifold hard projection, PCA ($^\ast$) & +13.2 & +5.8 & +2.6 & +12.2 & +0.1 & +5.3 & +7.2 & +3.4 & +6.7 & +3.4 \\
    ($^\ast$) + Fine-tuning \& distillation      & \textbf{+14.1} & +6.3 & \textbf{+2.8} & \textbf{+15.1} & \textbf{+0.9} & +5.7 & +9.3 & \textbf{+3.8} & +6.7 & +3.5 \\
    \midrule
    ($^\ast$) + Fine-tuning \& distillation      & 82.5 & 84.3 & 76.2 & 75.5 & 79.3 & 84.2 & 73.8 & 78.0 & 78.6 & 63.7 \\
    ~ & \pp{2.4}~ & \pp{2.1}~ & \pp{0.4}~ & \pp{2.1}~ & \pp{0.4}~ & \pp{0.6}~ & \pp{2.5}~ & \pp{1.8}~ & \pp{1.0}~ & \pp{0.4}~ \\
    \bottomrule
  \end{tabularx}
\end{table}

\section{Additional results: VQA}
\label{appendixVqaResults}

The experiments of \secref{secVqa} use only the binary (yes/no) question of the GQA dataset~\cite{hudson2018gqa}.
We repeated them with the full dataset.
The setup, model, and hyperparameters are unchanged.
The models' accuracy remains close to that of the baseline (\tabref{tabVqaFull}) even when the independence constraint induces the models to focus on different input features.
In \figref{figVqaExamples}, we include examples from the validation set that illustrate this effect.
Contrary to \secref{secVqa}, we did not observe improvements in accuracy when training three models with our method.
Our ongoing work is exploring the range of hyperparameters (larger number of models, different regularizer weights) to further investigate these observations.

\begin{table}[h!]
  \caption{Experiments on the full GQA dataset. Models maintain very similar accuracy across the board while focusing on different image features.\label{tabVqaFull}}
  \vspace{4pt}
  \scriptsize
  \renewcommand{\tabcolsep}{0.15em}
  \renewcommand{\arraystretch}{1.1}
  \centering
  \begin{tabularx}{\linewidth}{Xcccccc}
  \toprule
  \textbf{GQA full} & N. of & GQA Val. & GQA Val. & GQA-OOD & GQA-OOD & Grad-CAM\\
  (accuracy in \%) & models & \scriptsize (best) & \scriptsize(ensemble) & Val-head \scriptsize (best) & Val-tail \scriptsize (best) & rank corr.\\
  \midrule
  Baseline &  3 & 49.7 & 52.3 & 53.3 & \textbf{35.6} & 0.1404 \\
  \midrule
  \multicolumn{5}{l}{With PCA manifold model}\\
  + Independence &  3 & 50.9 & \textbf{52.7} & 54.4 & 35.0 & 0.2230 \\
  + Ind. + on-manifold &  3 & \textbf{51.0} & 52.5 & 54.7 & 35.2 & 0.1868\\
  \midrule
  \multicolumn{5}{l}{With PCA manifold model}\\
  + Independence &  3 & 50.7 & 51.7 & 55.1 & 34.0 & 0.1626 \\
  + Ind. + on-manifold &  3 & 50.7 & 51.6 & \textbf{55.8} & 33.8 & \textbf{0.1012} \\
  \bottomrule
  \end{tabularx}
\end{table}

\begin{figure}[hb!]
  \centering
  \includegraphics[width=.95\linewidth]{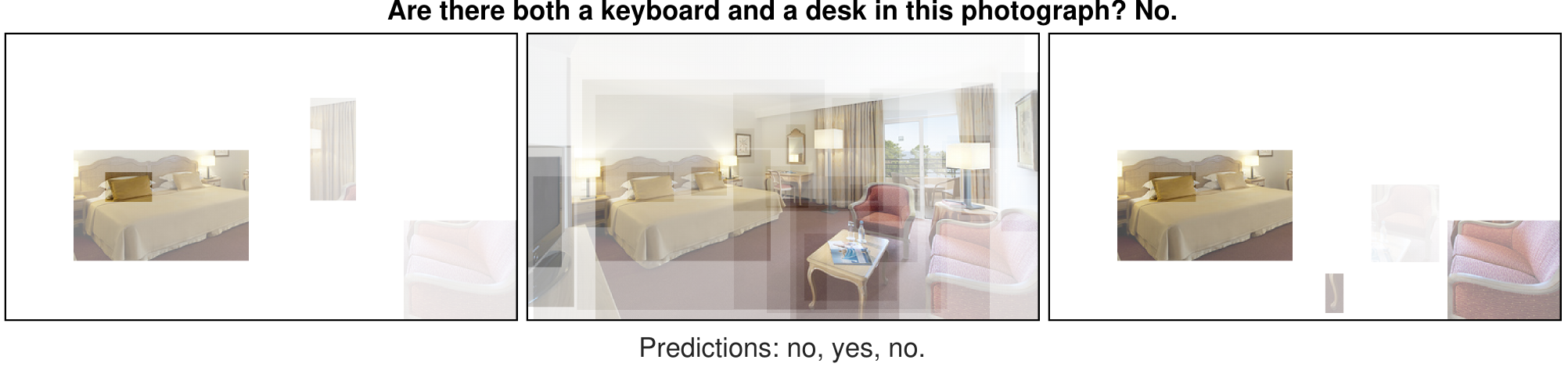}\vspace{14pt}\\
  \includegraphics[width=.95\linewidth]{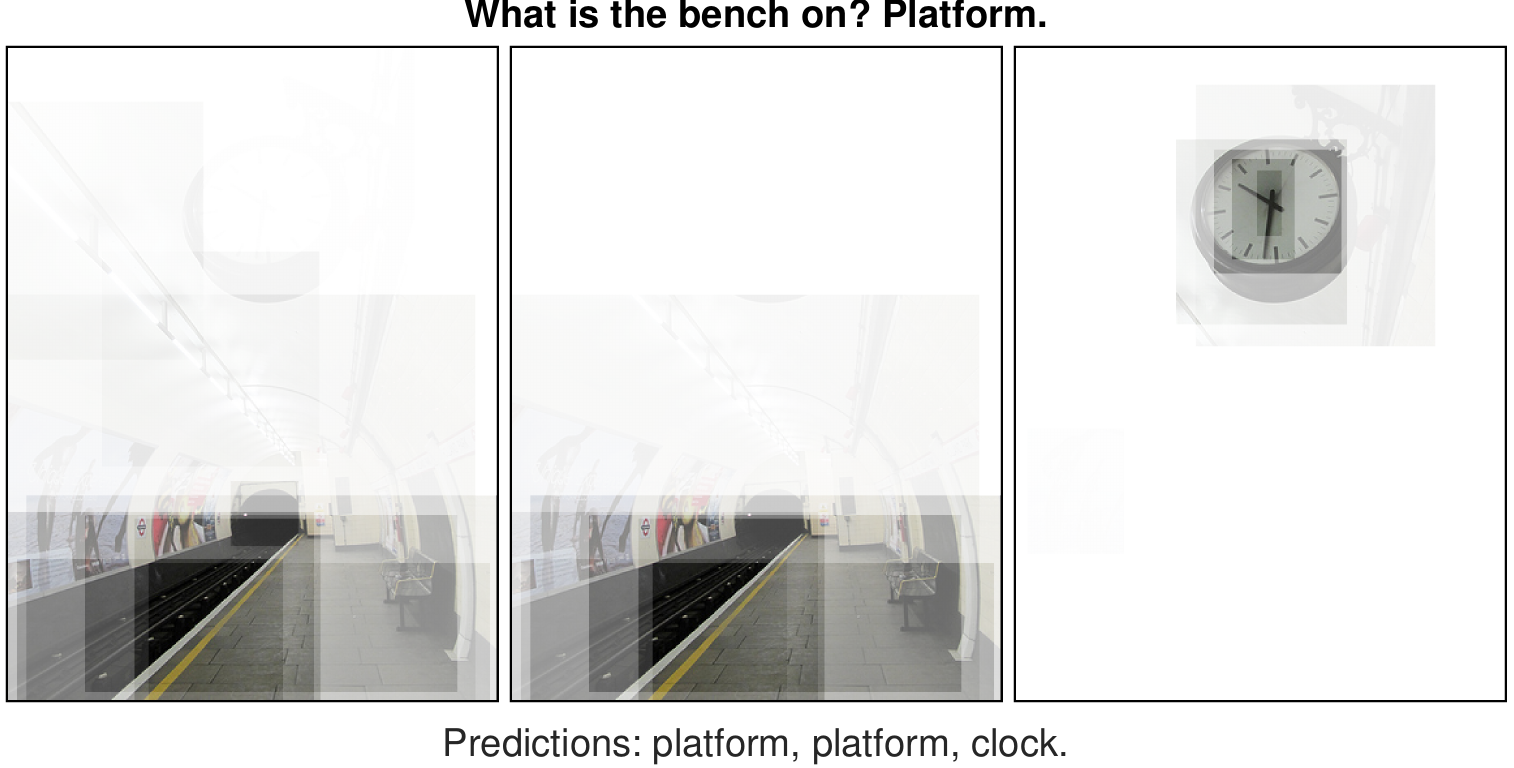}\vspace{14pt}\\
  \includegraphics[width=.95\linewidth]{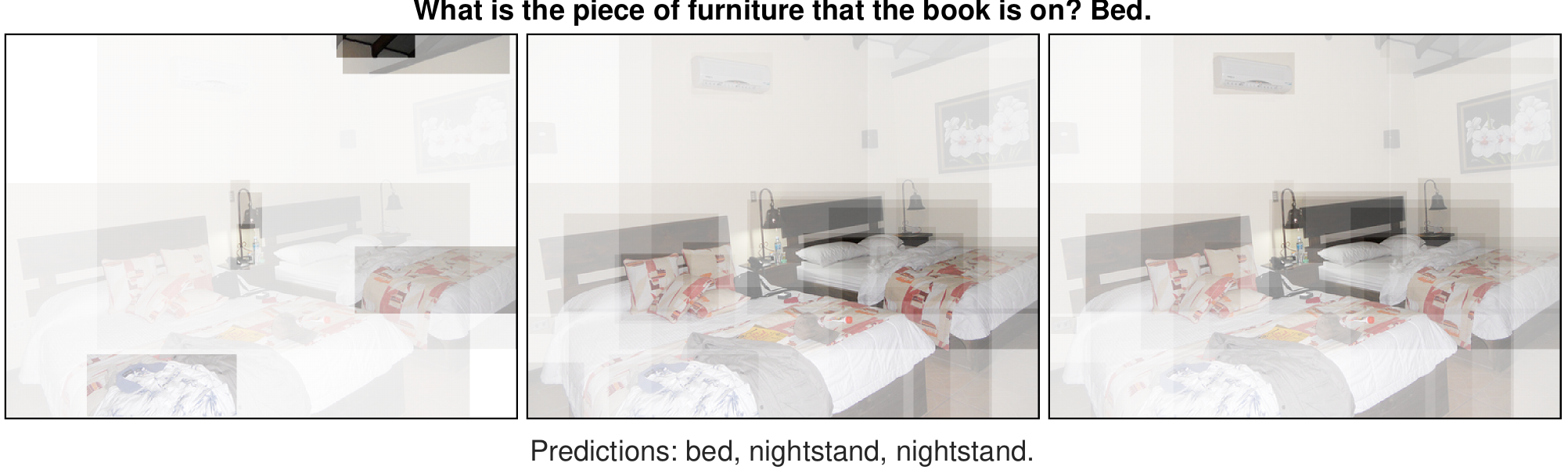}\vspace{4pt}\\
  \caption{
  Examples from the GQA validation set. We show the input question, ground truth answer, and predicted answer from 3 models trained with our method. The input images are weighted with grad-CAM scores over object detections.
  The models often predict the same answer while focusing on different regions.
  This diversity in spatial locations emerges naturally: our independence constraint is applied across the \emph{channels} of globally-pooled visual features.
  \label{figVqaExamples}}
\end{figure}




\end{document}